\PassOptionsToPackage{sort, compress}{natbib}

\documentclass[11pt, a4paper, colorlinks=true,allcolors=blue,hidelinks]{include/gdm_format}
\usepackage[numbers, sort&compress]{natbib}

\usepackage{amsmath,amsfonts,bm}

\def\eqref#1{equation~\ref{#1}}

\def\1{\bm{1}}

\DeclareMathAlphabet{\mathsfit}{\encodingdefault}{\sfdefault}{m}{sl}
\SetMathAlphabet{\mathsfit}{bold}{\encodingdefault}{\sfdefault}{bx}{n}

\usepackage{url}
\usepackage{booktabs} 
\usepackage{colortbl} 
\usepackage{xcolor} 
\usepackage{array} 
\usepackage{multirow} 
\usepackage{amssymb}
\usepackage{graphicx} 
\usepackage{float}    
\usepackage[most]{tcolorbox}
\usepackage{listings}
\usepackage{fontawesome5}
\usepackage{pgffor}
\usepackage{pgfplots}

\usepackage{amsmath}
\usepackage{amssymb} 
\usepackage{xspace}
\usepackage{multirow}
\usepackage{booktabs}
\usepackage{color}
\usepackage{colortbl}
\usepackage{cleveref}
\usepackage{graphicx}
\usepackage{algorithm}
\usepackage{algorithmicx}
\usepackage{algpseudocode}
\usepackage{subcaption}
\usepackage{wrapfig}
\usepackage{sidecap}
\usepackage{soul}
\usepackage{enumitem}
\sidecaptionvpos{figure}{t}
\usepackage{multicol}
\usepackage{pifont}
\usepackage[normalem]{ulem}
\usepackage{titletoc}
\usepackage{CJKutf8}
\usepackage{graphicx}
\usepackage{caption}
\usepackage{ulem}
\usepackage{nameref}
\newcolumntype{L}[1]{>{\raggedright\let\newline\\\arraybackslash\hspace{0pt}}m{#1}}
\newcolumntype{C}[1]{>{\centering\let\newline\\\arraybackslash\hspace{0pt}}m{#1}}
\newcolumntype{R}[1]{>{\raggedleft\let\newline\\\arraybackslash\hspace{0pt}}m{#1}}
\usepackage{tabularray}
\usepackage[export]{adjustbox}

\def\scititle{
SRT-H: A Hierarchical Framework for Autonomous Surgery via Language-Conditioned Imitation Learning}
\title{\bfseries \boldmath \scititle}
\usepackage{tikz}
\usetikzlibrary{fit}
\usetikzlibrary{positioning} 
\usetikzlibrary{calc}
\usepackage{pgfplots}
\pgfplotsset{compat=1.16}
\usepgfplotslibrary{fillbetween}
\usetikzlibrary{patterns}
\usepackage{siunitx}
\usepackage{pgf-pie}
\definecolor{OIblack}{RGB}{0, 0, 0}
\definecolor{OIgreen}{RGB}{0, 158, 115}
\definecolor{OIblue}{RGB}{0, 114, 178}
\definecolor{OIlightblue}{RGB}{86, 180, 233}
\definecolor{OIyellow}{RGB}{240, 228, 66}
\definecolor{OIorange}{RGB}{230, 159, 0}
\definecolor{OIred}{RGB}{213, 94, 0}
\definecolor{OIpink}{RGB}{204, 121, 167}
\usepackage{xspace}
\newcommand*{\eg}{\emph{e.g.}\@\xspace}

\usepackage[acronym]{glossaries}
\glsdisablehyper
\newacronym{ras}{RAS}{robot-assisted surgery}
\newacronym{loa}{LoA}{level of autonomy}
\newacronym{srt}{SRT}{surgical robot transformer}
\newacronym{vlm}{VLM}{Vision-Language Model}
\newacronym{hl}{HL}{high-level}
\newacronym{ll}{LL}{low-level}
\newacronym{ce}{CE}{cross-entropy}
\newacronym{ood}{OOD}{Out-of-Distribution}
\newacronym{gui}{GUI}{graphical user interface}
\newacronym{mlp}{MLP}{multi-layer perceptron}
\newacronym{dvrk}{dVRK}{da Vinci Research Kit}
\newacronym{dagger}{DAgger}{Dataset Aggregation}
\newacronym{cvs}{CVS}{critical view of safety}
\newacronym{fps}{FPS}{frames per second}

\lstdefinestyle{pythonstyle}{
    language=Python,
    basicstyle=\ttfamily\small,
    keywordstyle=\color{blue},
    commentstyle=\color{green!50!black},
    stringstyle=\color{red},
    showstringspaces=false,
    numbers=left,
    numberstyle=\tiny\color{gray},
    frame=single,
    breaklines=true,
    tabsize=4,
}

\tcbset {
  base/.style={
    arc=0mm, 
    bottomtitle=-0.25mm,
    boxrule=0mm,
    colbacktitle=black!10!white, 
    coltitle=black, 
    fonttitle=\bfseries, 
    left=2.5mm,
    leftrule=1mm,
    right=3.5mm,
    title={#1},
    toptitle=0.25mm,
    breakable,
  }
}

\definecolor{brandblue}{rgb}{0.34, 0.7, 1}
\newtcolorbox{mybox}[1]{
  colframe=brandblue, 
  base={#1}
}

\definecolor{pink}{rgb}{1, 0.75, 0.8}
\newtcolorbox{safetybox}[1]{
  colframe=pink, 
  base={#1}
}

\title{SRT-H: A Hierarchical Framework for Autonomous Surgery via Language-Conditioned Imitation Learning}

\correspondingauthor{Ji Woong (Brian) Kim (jwbkim@stanford.edu)}

\author[1,2]{Ji Woong (Brian) Kim}
\author[1]{Juo-Tung Chen}
\author[1]{Pascal Hansen}
\author[2]{Lucy X. Shi}
\author[1]{Antony Goldenberg}
\author[1]{Samuel Schmidgall}
\author[1]{Paul Maria Scheikl}
\author[1]{Anton Deguet}
\author[1]{Brandon M. White}
\author[3]{De Ru Tsai}
\author[3]{Richard Cha}
\author[1]{Jeffrey Jopling}
\author[2]{Chelsea Finn}
\author[1]{Axel Krieger}

\affil[1]{Johns Hopkins University}
\affil[2]{Stanford University}
\affil[3]{Optosurgical}

\begin{document}

\begin{abstract}

Research on autonomous surgery has largely focused on simple task automation in controlled environments.  However, real-world surgical applications demand dexterous manipulation over extended durations and generalization to the inherent variability of human tissue. These challenges remain difficult to address using existing logic-based or conventional end-to-end learning approaches. To address this gap, we propose a hierarchical framework for performing dexterous, long-horizon surgical steps. Our approach utilizes a high-level policy for task planning and a low-level policy for generating robot trajectories. The high-level planner plans in language space, generating task-level or corrective instructions that guide the robot through the long-horizon steps and correct for the low-level policy's errors. We validate our framework through ex vivo experiments on cholecystectomy, a commonly-practiced minimally invasive procedure, and conduct ablation studies to evaluate key components of the system. Our method achieves a 100\% success rate across eight unseen ex vivo gallbladders, operating fully autonomously without human intervention. This work demonstrates step-level autonomy in a surgical procedure, marking a milestone toward clinical deployment of autonomous surgical systems.

\end{abstract}

\maketitle

\begin{center}
\href{https://h-surgical-robot-transformer.github.io/}{\faGithub  \xspace \texttt{https://h-surgical-robot-transformer.github.io/}}
\end{center}

\section{Introduction}

Autonomous surgery offer the potential to improve surgical outcomes, reduce costs, and expand access to high-quality care.  However, most surgical robots today remain teleoperated due to fundamental challenges. From a vision perspective, surgical scenes are highly complex, involving morphological variation between patients, constant environmental changes during interventions, and visual occlusions such as blood and smoke from cautery tools. Motion planning in this setting is difficult, because of the partial observability of organs and their unpredictable dynamics. Additionally, surgical tasks must be performed with high precision and safety, making the development of these systems very challenging.

Prior works have addressed surgical autonomy through various strategies in simulation~\cite{scheikl2023lapgym, yu2024orbit, xu2021surrol} and real-world settings~\cite{saeidi2022autonomous, liang2024medic, fagogenis2019autonomous, thananjeyan2017patterncutting, kuntz2023autonomous}. Various studies explored tabletop tasks such as peg transfer, needle pickup, and deformable object manipulation, using model-based strategies~\cite{hwang2022automating, afshar2022tissuemanipulation, liang2024medic, fagogenis2019autonomous, hu2024tissue}, reinforcement learning~\cite{ou2023sim2real, scheikl2023sim2real, chiu2021BimanualRegrasping, thananjeyan2017patterncutting, xu2021surrol, haiderbhai2024sim2real}, and imitation learning~\cite{scheikl2024mpd, shin2019tissuemanipulation, tanwani2021sequential, su2021tbd, pore2021lfd}. In particular, learning-based methods show promise in tackling challenging contact-rich manipulation tasks~\cite{schmidgall2024general}, such as suture knot-tying~\cite{kim2024surgical}, which are otherwise difficult to solve with model-based strategies. Although promising, most learning-based works were demonstrated in controlled environments and have not been extended to realistic in-vivo or ex-vivo settings. Therefore, whether these strategies will succeed in the complex and diverse environment of surgery remains uncertain.

On the other hand, there have been notable in-vivo autonomous demonstrations such as needle steering~\cite{kuntz2023autonomous} and anastomosis tasks~\cite{saeidi2022autonomous}. Although promising, these studies primarily tackled the navigation steps of the procedure, which is much simpler than manipulation, and relied on hand-crafted strategies that were specifically optimized for a single application. In-vivo studies demonstrate the promise of robotics being deployed in clinically relevant environments, however, the applied strategies are unlikely to generalize, scale, or address complex manipulation problems which are very common in surgery.

  In this work, we aimed to move beyond the scope of prior approaches by addressing several critical and previously unaddressed dimensions of surgical autonomy. First, we focus on contact-rich manipulation tasks that require diverse tool use, including grabbing, clipping, and cutting. Second, we conduct this work in a realistic ex-vivo setting with significant variability in tissue appearance, anatomy, and morphology across organs, mirroring the diversity encountered in human surgeries. Third, rather than tackling individual skills, we tackle entire surgical steps that unfold over several minutes and require persistent coordination and decision-making. The combination of these challenges has been unexplored in prior work and is non-trivial to solve using conventional approaches. Our goal is to show that these challenges can be overcome with a unified design using data-drive methods. Solving this challenge in such generalizable way is essential for progressing toward clinically viable and general-purpose autonomous systems.
  

\begin{figure}[htbp]
    \captionsetup{labelformat=empty} 
    \centering
    \includegraphics[width=0.9\linewidth]{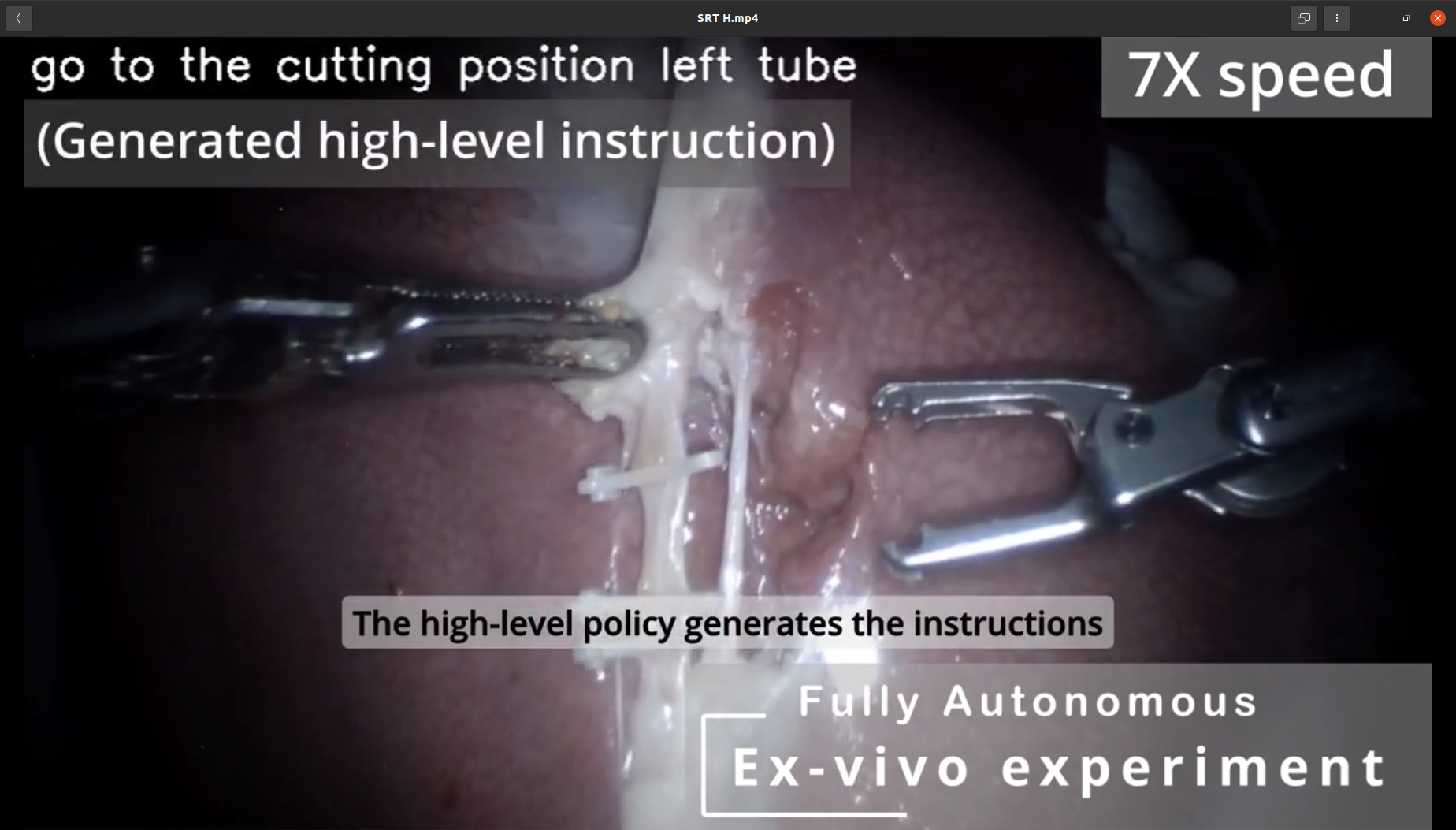}
    \caption{\textbf{\href{https://h-surgical-robot-transformer.github.io/resources/tasks_srt_h/overall_summary_vid_srt_h.mp4}{Movie 1}:} A comprehensive summary of our work. Using cholecystectomy as a case study, our framework automates key steps in gallbladder removal, focusing on the complex process of clipping and cutting the cystic duct and artery. The system performs 17 tasks fully autonomously, achieving successful results in all eight ex-vivo studies without human intervention. Robustness is demonstrated through challenging scenarios and appearance variations, where the model adapts and executes tasks confidently, highlighting its potential for generalizing across surgical settings.}
    \label{fig:setup}
\end{figure}

\setcounter{figure}{0}
\begin{figure}[htbp]
    \centering
    \includegraphics[width=0.9\linewidth]{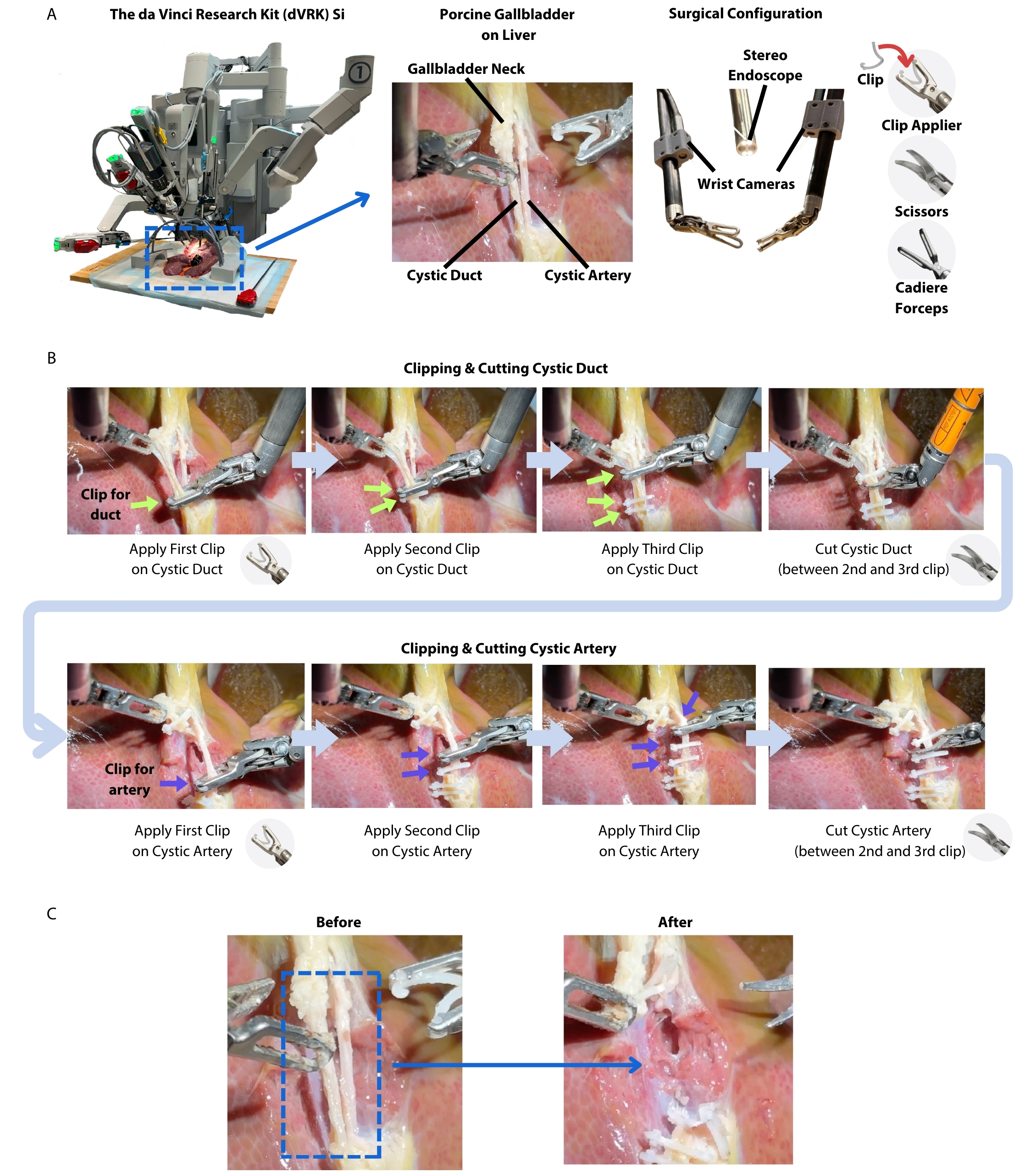}
    \caption{\textbf{System and task overview.} (\textbf{A}) We use the da Vinci Research Kit (dVRK) Si to deploy our policy, which includes an endoscope and two additional wrist cameras mounted for a better view of the interactions between instruments and tissue. (\textbf{B}) The autonomous surgical steps include clipping and cutting the gallbladder's artery and duct. (\textbf{C}) The before and after pictures illustrate the objective of this procedure; the duct and artery are completely severed, without spilling any of their internal fluids thanks to the use of clips.}
    \label{fig:setup}
\end{figure}


Towards this end, we present Hierarchical Surgical Robot Transformer (SRT-H), a framework for autonomous, step-level autonomy in surgery (Movie 1). SRT-H uses a hierarchical architecture composed of a high-level (HL) policy that issues natural language instructions, including task and corrective instructions, and a low-level (LL) policy that executes low-level trajectories. This structure allows us to decompose complex procedures into shorter tasks and enable the HL policy to correct mistakes made by the LL policy, which will naturally arise during long-horizon steps. Furthermore, using language enables an intuitive interface for intermittent user intervention and fine-tuning. Specifically, users can temporarily override HL decisions with natural language instructions, and these interventions are stored and used for continual learning via a DAgger-style loop \cite{dagger}.

SRT-H is built on a transformer-based architecture and trained end-to-end via imitation learning, using only red, green, blue (RGB) images paired with language annotations. It avoids reliance on depth sensors, segmentation modules, or specialized hardware. We evaluate SRT-H on the clipping-and-cutting step of cholecystectomy, a common laparoscopic procedure performed over 700,000 times annually in the United States \cite{acalovschi2012growing}. This step involves identifying the cystic duct and artery, placing clips, and severing them. By disabling the clip latching mechanism, we enable collection of hundreds of demonstrations from a single porcine tissue, making large-scale data collection feasible. In contrast, other steps like dissection are destructive and yield only one demonstration per specimen, motivating our focus on clipping and cutting steps of cholecystectomy.

To train and evaluate our system, we collect 16,000 trajectories (approximately 17 hours of data) across 34 ex-vivo porcine gallbladders. We then test SRT-H on eight unseen gallbladders, and in each case, the system successfully completed all 17 required tasks autonomously, generalizing across anatomies and self-correcting its mistakes mid-procedure. Ablation studies highlight the critical role of both the hierarchical structure and the corrective language interface in enabling timely and effective corrective behaviors. Compared to an expert surgeon, our framework shows comparable performance, but requries longer execution time. In summary, SRT-H provides a scalable and adaptable framework for autonomous surgery, with potential to advance toward generalizable autonomy in real-world surgical settings and further in vivo studies.

\begin{figure}[htbp]
    \centering
    \includegraphics[width=0.9 \linewidth]{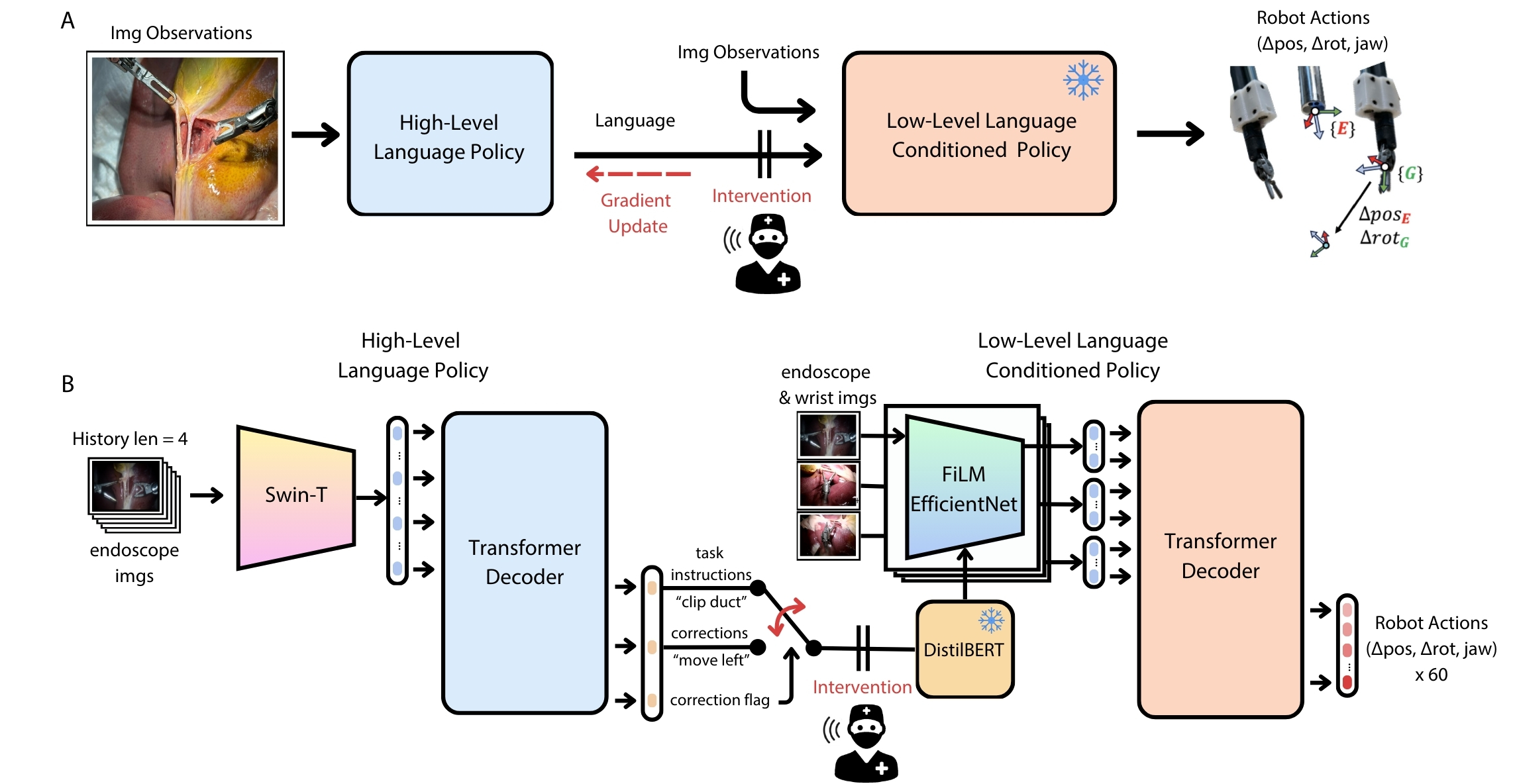}
    \caption{\textbf{Model overview and architecture.} \textbf{(A)} The architecture of our framework consists of a high-level policy that generates language instructions given the image observations, and a low-level policy that conditions on the language instructions and image observations to generate robot motions in Cartesian space. \textbf{(B)} On a more granular level, the high-level policy consists of a Swin-T model to encode the visual observations into tokens, that are processed by a Transformer Decoder to generate language instructions. The language instructions are processed by a pretrained and frozen distilled bidirectional encoder representations from transformers (DistilBERT) model to generate language embeddings. The image observations are passed to an EfficientNet that conditions on the language embeddings through feature-wise linear modulation (FiLM) layers. The combined embeddings are passed to a Transformer Decoder to generate a sequence of actions that are encoded in delta position and orientation values.}
    \label{fig:hl_ll}
\end{figure}

\section*{\textbf{RESULTS}}
In the following sections, we describe the design and workflow of our autonomous surgery system and then present the experiment results. We first evaluate our system's ability to complete the cholecystectomy procedures using eight unseen ex-vivo porcine tissues. The framework's performance was evaluated based on the success rate, total time, and number of self-corrections made (see ``Core experiment results" section). We further evaluated SRT-H against ablative variants to show the effect of different design choices on the performance of the framework. We evaluated these variants based on their success rate, total time, and ability to recover from failure states (see ``Comparison with variants" section). The success rate of failure recoveries were evaluated by placing the instruments into failure states and observing whether each variant can recover to complete the procedure successfully. We also independently performed ablative comparisons for the \gls{hl} policy and quantified each design choice's effect on its performance (see ``High-level policy ablative studies" section). Lastly, we evaluated our framework against an expert surgeon based on the success rate, time to completion, and the smoothness of the trajectories (see ``Comparison with expert surgeon" section).

\subsection*{\textbf{Experiment design}}

Figure~\ref{fig:setup}A shows the hardware configuration of our system, which consists of a da Vinci Research Kit (dVRK) Si with wrist cameras mounted near the instrument tips. The stereo endoscope of the \gls{dvrk} provides a global view of the surgical scene, and the wrist cameras provide a close-up view of interactions between instruments and tissue. Prior works~\cite{hsu2022visionbased, kim2024surgical} demonstrated that wrist cameras can help with generalizing to different workspace heights and out-of-distribution scenarios due to the more consistent view provided by the wrist cameras. Though the size of the wrist cameras used in this study are quite large and perhaps not clinically practical for minimally invasive surgery, their design can be further downsized. 

In the following, we describe the general workflow of the procedure within cholecystectomy that is automated, its challenges, and the steps for deploying SRT-H. The steps for clipping and cutting the duct and artery are shown in Fig.~\ref{fig:setup}B. The objective of this step is as follows: three clips are added to the left tubular structure (typically the duct) and then three clips are added to the right tubular structure (typically the artery). For each tube, the first two clips are placed proximally near the bottom and the third clip distally at the top. Note that the clips prevent any leakage of biological fluids after the gallbladder is removed; in particular, the two clips placed at the base remain in the patient and must therefore provide a secure, long-lasting seal.
Then, the tube is transected between the second and third clip of each tube, where
there is the most gap for the scissors to enter. In general, the duct and artery are in close proximity,
therefore, the left gripper must apply tension at the neck of the gallbladder to stretch the tubes apart
and make room for the clip applier or the scissor to enter the gap. After each clip is applied, an
assistant on standby near the dVRK loads another clip and also performs tool changes between clip
applier and scissors after completion of the relevant steps (filling the role of a surgical nurse).

There are several challenging elements to this procedure. From a visual and anatomical point of view, the appearance of the ducts and arteries vary greatly between patients in terms of their diameter, length, proximity, angle from each other, and the amount of connective tissue left on the surface of the tubes, which can make perception challenging \cite{gupta2023anatomical}. From a manipulation point of view, a precise bimanual coordination of the arms is necessary. In particular, when adding clips to the left tube, the left gripper must grab the neck of the gallbladder head and stretch it to make sufficient space between the duct and the artery, and the clip applier must pry in between the tight space between the tubes to successfully apply the clip \cite{majumder2020laparoscopic}. During this step, the clip applier can overshoot and miss the duct entirely, mistakenly clip the right tube (artery), or apply the clips at a suboptimal location e.g., applying the third clip too close to the second clip so as to leave no space for the scissors to perform the cut. Overall, to succeed in these steps, the policy must perceive and track the location of the deformable duct and artery, keep an internal count of how many clips have been applied so far, detect whether sufficient stretch has been applied to make room for prying in the clip applier tool, and apply the clips at an optimal location without damaging the surrounding tissues.

During the autonomous trials when SRT-H is deployed, the operator clicks a button on the \gls{gui} to initiate the system. After the system autonomously applies each clip, the system automatically pauses on its own and waits for the operator to load another clip. The operator then loads another clip and the procedure is resumed. This interaction is repeated for all six clips that are applied to the duct and artery. Between the clip-applying steps, when a scissor is required, similar steps are carried out; the robot autonomously requests for a tool change, and the operator resumes the procedure after making the tool change.

The architectural details of SRT-H are shown in Fig.~\ref{fig:hl_ll}. Briefly, SRT-H is implemented as two transformer decoders, one is part of the \gls{hl} policy and the other of the \gls{ll} policy. The \gls{hl} policy takes in a history of endoscope images as input and generates three outputs, which includes the task instruction, corrective instructions, and correction flag (boolean). Either the task or corrective instruction is provided as input to the \gls{ll} policy, with the correction flag serving as a binary switch that determines which instruction is sent to the \gls{ll} policy. The \gls{ll} policy then takes the given instruction, along with the current observations of the surgical scene, to generate a hybrid-relative trajectory~\cite{kim2024surgical}, the action representation optimized for training on \gls{dvrk} robots.

\subsection*{\textbf{Core experiment results}}

\begin{table}[htbp]
\caption{\textbf{Core Experiment Metrics.} Procedures were performed on n=8 ex-vivo porcine gallbladder tissues; the metrics include success rates, total duration, and number of self-corrections over all tasks of the procedure.}
\label{tab:experiment_metrics}
\centering
\begin{tabular}{l c c c}
\toprule
 & \multicolumn{1}{C{2cm}}{\textbf{Success Rate (\%)}} & \textbf{Duration (s)} & \multicolumn{1}{C{2.2cm}}{\textbf{\# Self-Corrections}} \\
 \midrule
\textbf{Gallbl. 1} &  100 & 290 &  2 \\
\textbf{Gallbl. 2} &  100 & 315 &  8 \\
\textbf{Gallbl. 3} &  100 & 304 & 14 \\
\textbf{Gallbl. 4} &  100 & 300 &  3 \\
\textbf{Gallbl. 5} &  100 & 396 &  6 \\
\textbf{Gallbl. 6} &  100 & 318 & 12 \\
\textbf{Gallbl. 7} &  100 & 274 &  1 \\
\textbf{Gallbl. 8} &  100 & 337 &  5 \\
\midrule
\textbf{Average}   &  100 & 317 & 6 \\
\bottomrule
\end{tabular}
\end{table}

\begin{figure}[htbp]
    \centering
    \includegraphics[width=0.9 \linewidth]{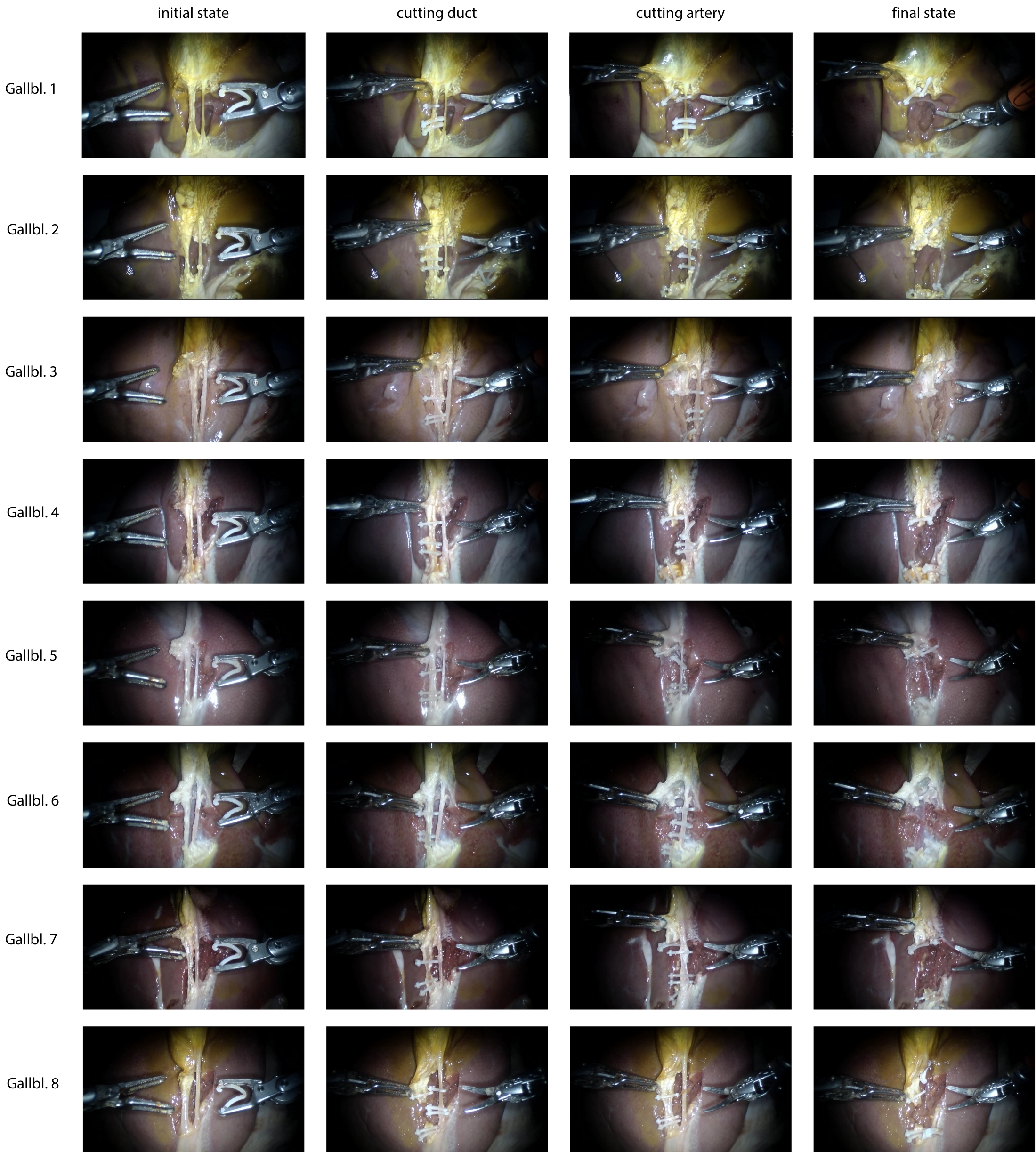}
    \caption{\textbf{Core experiment sequences.} Images of the initial and final states, as well as observations of the clip positions for the duct and artery before the cut is made for all eight gallbladders. The clips are sufficiently secured around the ducts and arteries, and sufficient space between the second and third clips of each tube is left for the scissors to make the cuts. The individual gallbladders vary noticeably in color, texture, and anatomy.}
    \label{fig:core_exp_sequence}
\end{figure}

For the core experiments, SRT-H was evaluated on eight different unseen gallbladders.
Table~\ref{tab:experiment_metrics} shows the result of each experiment including the success rate, total duration, and number of self-corrections made.
We observe that SRT-H was able to complete all the procedures successfully without any human interventions, and on average completed the procedure within 317 seconds or 5 minutes and 17 seconds. This duration excludes the time of reloading the clips and making tool changes performed by the operator. Furthermore, when failure states were encountered, SRT-H was able to correct its own mistakes and complete the procedure successfully. On average, the self-corrections were made approximately six times throughout the entire procedure.
We provide additional information about the individual self-corrections in Fig.~\ref{fig:selfcorrections}.
Figure~\ref{fig:core_exp_sequence} shows the placement of each clip before the artery and duct were cut in more detail. It can be observed that the clips fully encompassed the ducts and arteries, maintained close spacing between the bottom two clips on each tube, and left sufficient spacing between the second and third clip on each tube for easy access for the scissors to make the cut. Overall, across diverse tissues, SRT-H demonstrated consistent capability in recognizing the relevant tissue structures, maintaining a reasonable pace, and recovering itself from its own failures to complete all cases successfully. In general, the upper-most clips were placed close to the gallbladder infundibulum, but at times, they may not have been positioned at the highest point. To alleviate such placement issues in the future, we may collect additional data where clips are positioned as far up as possible, allowing SRT-H to more accurately replicate ideal placement. Similarly, in some cases, the clips were placed quite low in the surgical field due to suboptimal demonstrations, and these issues could similarly be improved by collecting better demonstrations.

Additionally, we encountered a non-safety critical robot failure in one of the eight experiments that was not related to SRT-H, when the scissors broke and had to be replaced before continuing.
In addition, the \gls{dvrk} system had to be reinitialized three times during manual tool changes, also unrelated to SRT-H. Note that these issues arose because we were using the very first \gls{dvrk} Si still undergoing development, and the hardware system was not yet perfected. These hardware-related issues have since been resolved.
\begin{figure}[htbp]
    \begin{tikzpicture}
        \begin{axis}[
            ybar,
            nodes near coords={
                \pgfmathprintnumber[zerofill, fixed, precision=1]{\pgfplotspointmeta}
            },
            every node near coord/.append style={font=\footnotesize, text=black, rotate=90, anchor=east},
            name=SuccVsRec,
            title={Subtask and Recovery Success Rate (n=3)},
            symbolic x coords={ours, task only, no wrist, no HL dagger, end to end},
            xticklabels={\textbf{SRT-H (ours)}, task only, no wrist, no HL DAgger, end to end},
            xtick=data,
            ymin=0,
            ymax=110,
            xlabel={},
            ylabel={Success Rate (\%)},
            y label style={at={(-0.08, 0.5)}},
            width=0.5\textwidth,
            height=0.5\textwidth,
            x tick label style={
                rotate=45,
                anchor=east,
                inner sep=0pt,
                outer sep=1mm,
                align=center,
                font=\footnotesize,
            },
            y tick label style={
                font=\footnotesize,
            },
            xtick pos=bottom,
            bar width=4mm,
            ymajorgrids=true,
            yminorgrids=true,
            minor y tick num=2,
            major grid style={thick, black!30!white, dashed},
            minor grid style={ultra thin, black!20!white, dashed},
            major x tick style = transparent,
            minor y tick style = transparent,
            enlarge x limits=0.1,
            legend style={
                legend columns=2,
                font=\small,
                /tikz/every even column/.append style={column sep=1.5mm},
            },
            legend image code/.code={
                \draw [#1, draw=none] (0mm,-1mm) rectangle (1.5mm,2mm); },
            ]

            \addplot+ [
                fill=OIblue,
                draw=none,
            ] coordinates {
                (ours, 100)
                (task only, 100)
                (no wrist, 77.77766667)
                (no HL dagger, 77.77766667)
                (end to end, 33.332222)
            };
            \addlegendentry{Subtask};
            
            \addplot+ [
                fill=OIlightblue,
                draw=none,
            ] coordinates {
                (ours, 100)
                (task only, 66.66666667)
                (no wrist, 50)
                (no HL dagger, 75)
                (end to end, 33.333333)
            };
            \addlegendentry{Recovery};
        
        \end{axis}
        \begin{axis}[
            name=SuccVsData,
            at=(SuccVsRec.right of south east),
            anchor=left of south west,
            title={Success Rate vs Amount of Training Data (n=3)},
            nodes near coords={
                \pgfmathprintnumber[zerofill, fixed, precision=1]{\pgfplotspointmeta}
            },
            every node near coord/.append style={font=\footnotesize, outer sep=1mm, text=black},
            ymin=0,
            ymax=110,
            xlabel={},
            ylabel={Success Rate (\%)},
            y label style={at={(-0.08, 0.5)}},
            xlabel={Amount of Training Data},
            width=0.5\textwidth,
            height=0.5\textwidth,
            xtick=data,
            xticklabels={33.3\%, 66.6\%, 100.0\%},
            xmin=20,
            xmax=110,
            y tick label style={
                font=\footnotesize,
            },
            xtick pos=bottom,
            bar width=\mybarwidth,
            ymajorgrids=true,
            yminorgrids=true,
            minor y tick num=2,
            major grid style={thick, black!30!white, dashed},
            minor grid style={ultra thin, black!20!white, dashed},
            minor y tick style = transparent,
            legend style={
                legend columns=2,
                font=\small,
            },
            legend image code/.code={
                \draw [#1, draw=none] (0mm,-1mm) rectangle (1.5mm,2mm); },
            ]

            \addplot[color=OIblue, mark=square, ultra thick] coordinates {
                (33.333333,66.66666633)
                (66.666666,77.77777733)
                (100, 100)
            };        
            
        \end{axis}
        
        \makeatletter
        \newcommand\resetstackedplots{
        \makeatletter
        \pgfplots@stacked@isfirstplottrue
        \makeatother
        \addplot [forget plot,draw=none] coordinates{(ours,0) (taskonly,0) (nowrist,0) (nodagger,0) (endtoend,0)}; 
        }
        \makeatother

        \newcommand{\printcustvalue}{}
        \begin{axis}[
            name=TaskCompTime,
            at=(SuccVsRec.south west),
            yshift=-2.7cm,
            anchor=north west,
            ybar stacked,
            nodes near coords={\printcustvalue},
            nodes near coords greater equal only/.style={
                every node near coord/.append style={
                    check for match/.code={
                        \begingroup
                        \pgfkeys{/pgf/fpu}
                        \pgfmathparse{\pgfplotspointmeta>=#1}
                        \global\let\result=\pgfmathresult
                        \endgroup
                        \pgfmathfloatcreate{1}{1.0}{}
                        \let\ONE=\pgfmathresult
                        \ifx\result\ONE
                           \renewcommand{\printcustvalue}{\pgfmathprintnumber{\pgfplotspointmeta}*}
                        \else
                           \renewcommand{\printcustvalue}{\pgfmathprintnumber{\pgfplotspointmeta}}
                        \fi
                    },
                    check for match,
                },
            },
            nodes near coords greater equal only=90.0,
            every node near coord/.append style={font=\scriptsize, text=black},
            title={Average Completion Time (n=3)},
            symbolic x coords={ours, taskonly, nowrist, nodagger, endtoend},
            xticklabels={\textbf{SRT-H (ours)}, task only, no wrist, no HL DAgger, end to end},
            xtick=data,
            ymin=0,
            ymax=349,
            xlabel={},
            ylabel={Average Time (s)},
            width=0.995\textwidth,
            height=0.5\textwidth,
            x tick label style={
                rotate=45,
                anchor=east,
                inner sep=0pt,
                outer sep=1mm,
                align=center,
                font=\footnotesize,
                yshift=-2.0mm,
            },
            y tick label style={
                font=\footnotesize,
            },
            xtick pos=bottom,
            bar width=8mm,
            ymajorgrids=true,
            yminorgrids=true,
            minor y tick num=2,
            major grid style={thick, black!30!white, dashed},
            minor grid style={ultra thin, black!20!white, dashed},
            major x tick style = transparent,
            minor y tick style = transparent,
            enlarge x limits=0.1,
            legend style={
                legend columns=3,
                font=\small,
                /tikz/every even column/.append style={column sep=1.5mm},
            },
            legend pos=north west,
            legend image code/.code={
                \draw [#1, draw=none] (0mm,-1mm) rectangle (1.5mm,2mm);
                },
            ]
            
            \addplot+[bar shift=-5mm, ybar, black, fill=OIblue, draw=none] plot coordinates {
                (ours, 23)
                (taskonly, 20)
                (nowrist, 27)
                (nodagger, 25)
                (endtoend, 41)
            };
            \label{clip1}
            
            \addplot+[bar shift=-5mm, ybar, black, fill=OIblue!70, draw=none] plot coordinates {
                (ours, 46)
                (taskonly, 21)
                (nowrist, 47)
                (nodagger, 46)
                (endtoend, 90)
            };
            \label{clip2}
            
            \addplot+[bar shift=-5mm, ybar, black, fill=OIblue!40, draw=none] plot coordinates {
                (ours, 29)
                (taskonly, 27)
                (nowrist, 47)
                (nodagger, 42)
                (endtoend, 65)
            };
            \label{cut}
            
            \coordinate (outlierE) at (axis cs:endtoend, 125);

            \resetstackedplots
            
            \addplot+[bar shift=5mm, ybar, black, fill=OIgreen, draw=none] plot coordinates {
                (ours, 13)
                (taskonly, 18)
                (nowrist, 44)
                (nodagger, 15)
                (endtoend, 90)
            };
            \label{grasptop}
            
            \addplot+[bar shift=5mm, ybar, black, fill=OIgreen!70, draw=none] plot coordinates {
                (ours, 29)
                (taskonly, 42)
                (nowrist, 44)
                (nodagger, 25)
                (endtoend, 65)
            };
            \label{graspbottom}
            
            \addplot+[bar shift=5mm, ybar, black, fill=OIgreen!40, draw=none] plot coordinates {
                (ours, 21)
                (taskonly, 90)
                (nowrist, 66)
                (nodagger, 90)
                (endtoend, 90)
            };
            \label{clipcaught}
            
            \addplot+[bar shift=5mm, ybar, black, fill=OIgreen!20, draw=none] plot coordinates {
                (ours, 25)
                (taskonly, 23)
                (nowrist, 67)
                (nodagger, 24)
                (endtoend, 58)
            };
            \label{overshoot}
        \node [draw,fill=white, anchor=north west, outer sep=2mm] (legend1) at (rel axis cs: 0.0,1.0) {\shortstack[l]{
        Subtasks \\
        \ref{clip1} 1st clip \\
        \ref{clip2} 2nd clip \\
        \ref{cut} cut
        }};
        \node[xshift=-8mm] (helper1) at (axis cs:ours,110) {};
        \node[xshift=8mm] (helper2) at (axis cs:ours,110) {};

        \node[xshift=-8mm] (helper3) at (axis cs:taskonly,185) {};
        \node[xshift=8mm] (helper4) at (axis cs:taskonly,185) {};
        
        \node[xshift=-8mm] (helper5) at (axis cs:nowrist,230) {};
        \node[xshift=8mm] (helper6) at (axis cs:nowrist,230) {};

        \node[xshift=-8mm] (helper7) at (axis cs:nodagger,160) {};
        \node[xshift=8mm] (helper8) at (axis cs:nodagger,160) {};
        
        \node[xshift=-8mm] (helper9) at (axis cs:endtoend,310) {};
        \node[xshift=8mm] (helper10) at (axis cs:endtoend,310) {};
        \end{axis}
        \draw [line width=1.5pt, decorate,decoration={brace,amplitude=5pt}] (helper1) -- (helper2) node[font=\footnotesize, midway,yshift=5mm]{93.0};
        \draw [line width=1.5pt, decorate,decoration={brace,amplitude=5pt}] (helper3) -- (helper4) node[font=\footnotesize, midway,yshift=5mm]{120.5};
        \draw [line width=1.5pt, decorate,decoration={brace,amplitude=5pt}] (helper5) -- (helper6) node[font=\footnotesize, midway,yshift=5mm]{171.0};
        \draw [line width=1.5pt, decorate,decoration={brace,amplitude=5pt}] (helper7) -- (helper8) node[font=\footnotesize, midway,yshift=5mm]{133.5};
        \draw [line width=1.5pt, decorate,decoration={brace,amplitude=5pt}] (helper9) -- (helper10) node[font=\footnotesize, midway,yshift=5mm]{249.5};

        \node [draw,fill=white, anchor=north west, outer sep=2mm, xshift=-2mm] (legend2) at (legend1.north east) {\shortstack[l]{
        Recovery \\
        \ref{grasptop} grab top 
        \ref{graspbottom} grab bottom \\
        \ref{clipcaught} clip caught 
        \ref{overshoot} overshoot
        }};

        \node[anchor=east, xshift=-3mm] (A) at (SuccVsRec.north west) {\textbf{A}};
        \node[anchor=east, xshift=-3mm] (B) at (SuccVsData.north west) {\textbf{B}};
        \node[anchor=east, xshift=-3mm] (C) at (TaskCompTime.north west) {\textbf{C}};
    \end{tikzpicture}
    
    \caption{\textbf{Comparisons against variants.} \textbf{(A)} We compare the success rate of our method, SRT-H, against various variants on subtasks and recovery scenarios for n=3 gallbladders. These three gallbladders are independent of the eight gallbladders used in the experiment. \textbf{(B)} Shows the success rate of SRT-H for n=3 gallbladders with respect to the amount of training data used. \textbf{(C)} Shows the average completion time over n=3 gallbladders for SRT-H and ablative variants.}
    \label{fig:variant_comparisons}
\end{figure}
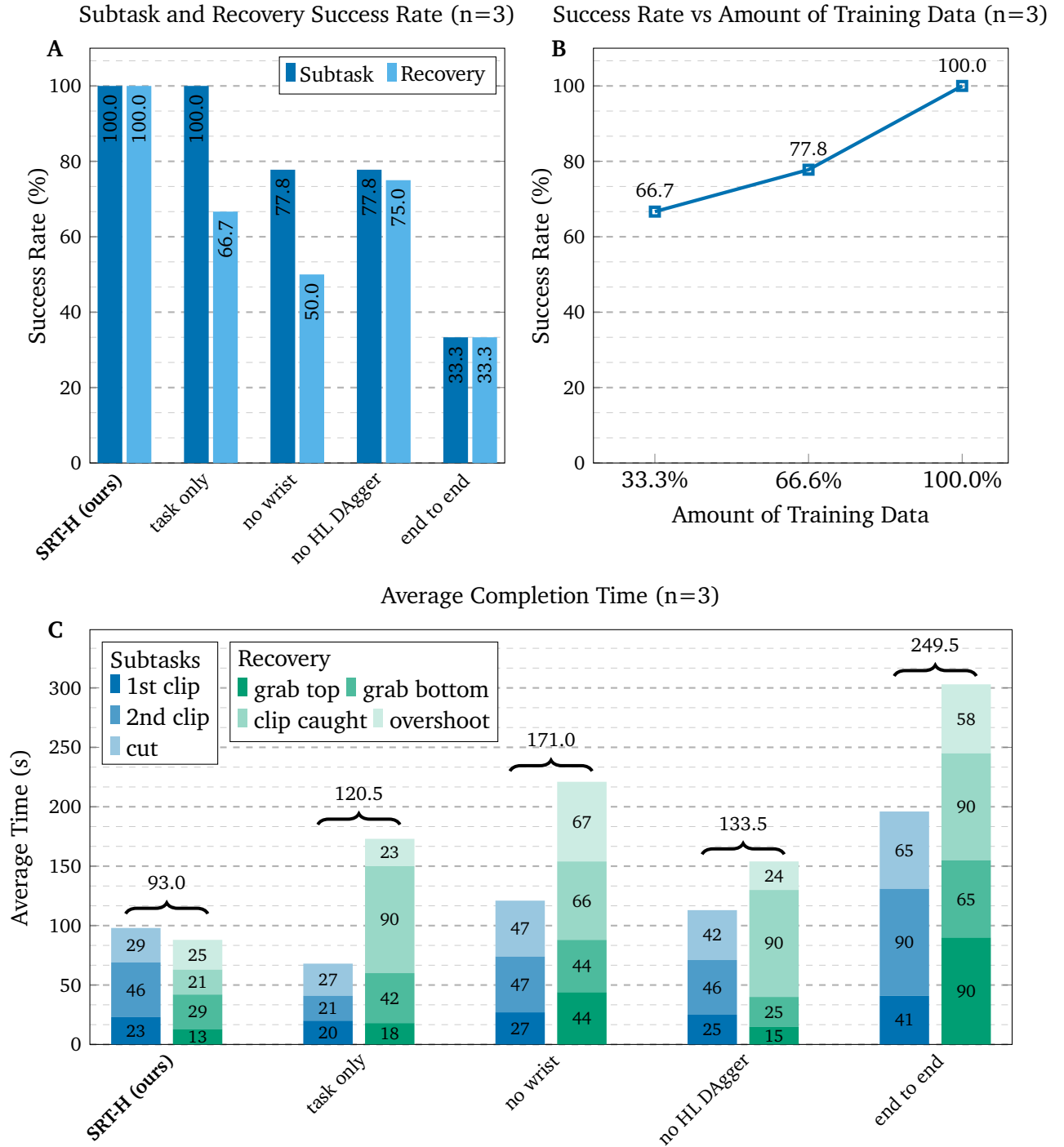

\begin{figure}[htbp]
    \centering
    \includegraphics[width=1\linewidth]{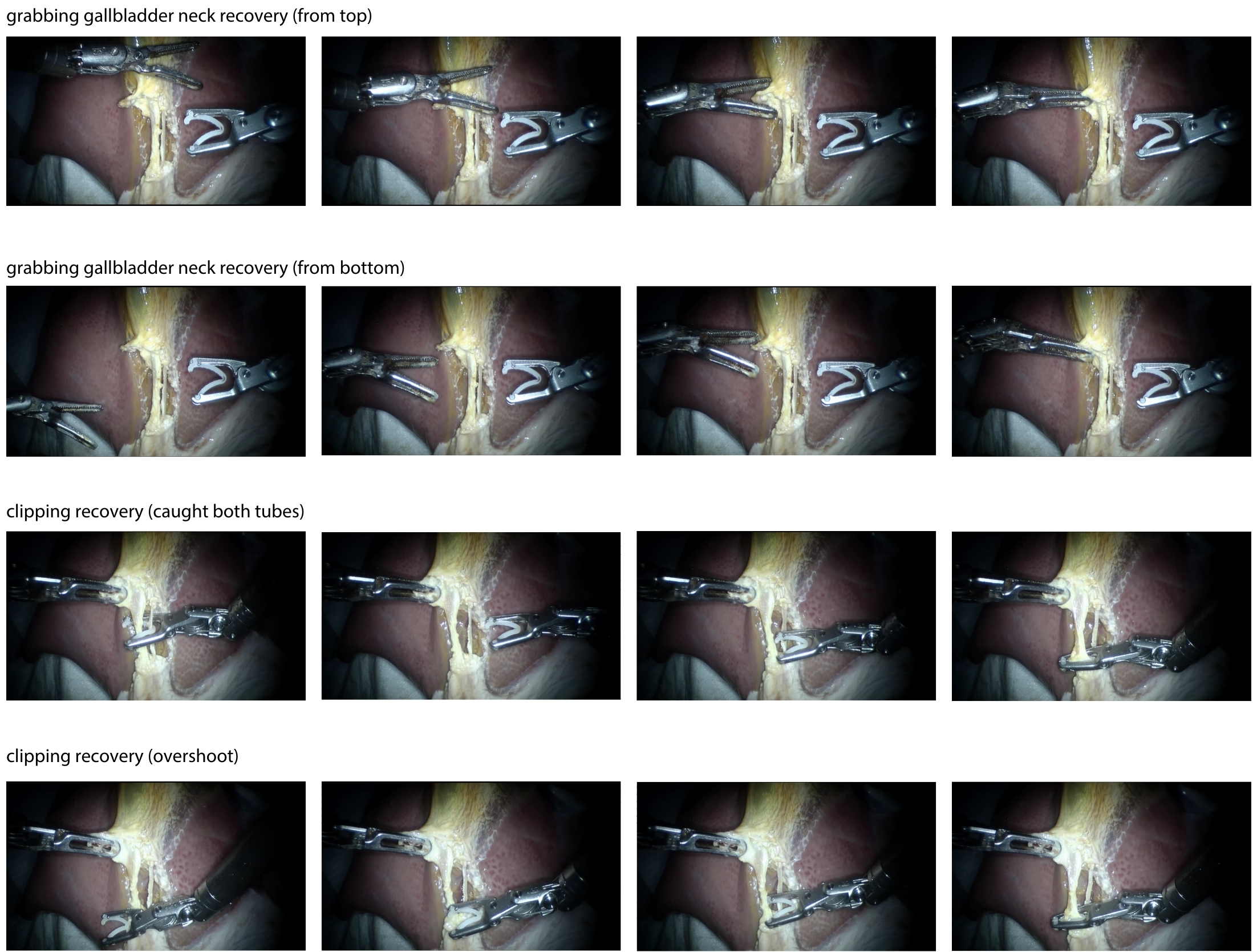}
    \caption{\textbf{Recovering from failure states.} We manually place the instruments into failure states to evaluate SRT-H's ability to recover from disadvantageous states of the environment. Each row illustrates a specific failure state and a sequence of images that show how SRT-H recovers from it.}
    \label{fig:recovery_test}
\end{figure}

\subsection*{\textbf{Comparison with variants}}
We further evaluated SRT-H against several variants, including SRT-H trained with task instructions only (no corrective instructions), SRT-H trained without wrist cameras, SRT-H's \gls{hl} policy trained without additional \gls{dagger} data (collected using expert language corrections during prior policy rollouts), and end-to-end architecture with only the \gls{ll} policy. For all the tests, each variant was evaluated based on its success rate and total duration. To ensure a fair comparison, all variants were evaluated using the same gallbladders and starting positions, with a \SI{90}{\s} maximum time limit set for completing each task. 


The full results of these evaluations are shown in Fig.~\ref{fig:variant_comparisons}. In terms of success rates (Fig.~\ref{fig:variant_comparisons}A), the results show that SRT-H scores the highest (100\%) in both normal and recovery scenarios (Fig. \ref{fig:recovery_test}). SRT-H using task instructions was a close second, as it also scored highest under normal scenarios (100\%), however, due to lack of corrective vocabulary, its performance in recovery scenarios was lower (66.7\%). Omitting wrist cameras also reduced the success rates in both scenarios (77.8\% and 50\% respectively), highlighting its importance in highly diverse ex-vivo scenarios beyond table-top settings. SRT-H without \gls{hl} fine-tuning resulted in diminished performance (77.8\% and 75\% respectively), demonstrating the importance of using a competent \gls{hl} policy and the efficacy of fine-tuning the \gls{hl} policy. The end-to-end policy variant scored the lowest in both scenarios (33.3\%).

In terms of total duration (Fig.~\ref{fig:variant_comparisons}C), results show that SRT-H performs the fastest on average for both normal and recovery scenarios. The other variants required more time due to making mistakes, which they could not recover from, or falling into repeating loops of retry behaviors. In general, however, the rate of motion for all variants was similar and their differences were dictated by how competent the policy was in recovery behaviors. 

We also evaluate how the amount of data affects policy performance. As shown in Fig.~\ref{fig:variant_comparisons}B, we evaluate SRT-H with 33.3\%, 66.6\%, and 100\% of the entire dataset as training data. These variants scored success rates of 66.7\%, 77.8\%, and 100\%, respectively. This evaluation indicates that beyond the design of the architecture, the amount of data plays a critical role in policy performance.

\subsection*{\textbf{High-level policy ablation studies}}
For the \gls{hl} policy, several design choices were made to address perception challenges arising from differences in gallbladder color, texture, and anatomy.
First, in addition to the full view, we incorporate a center-cropped version of the most critical operating area as input. The center-crop size is 432 $\times$ 480 pixels and the cropping location is always fixed on the original image.
This allows the model to focus on the most relevant information in the surgical field by providing this area at a higher resolution compared to the full view.
Second, we modify the cross-entropy-based loss function by scaling it with the $L_1$ distance between the predicted and reference task instructions.
This adjustment is intended to improve the policy’s ability to distinguish between tasks that are temporally distant but visually similar.
Third, to mitigate the effect of occlusions during surgery, we include a history of four past image frames, each spaced one second apart, along with the current frame.
This temporal context allows the \gls{hl} policy to retain crucial temporal information, ensuring robust performance even when important details are temporarily obscured.
We conduct an ablation study to determine the contribution of each design choice by systematically omitting each one during model training.
Performance is evaluated based on both accuracy and F1 score for three classification tasks: predicting task instructions, corrective instructions, and identifying recovery modes.

Results show that our \gls{hl} policy achieved an accuracy and F1 score of approximately 97\% for task instruction predictions. 
Removing the center crop input or using only the \gls{ce} loss for task instructions resulted in a decrease in accuracy and F1 score by around 2-2.5\%.
Omitting the observation history led to an even more substantial drop in performance, exceeding 10\% for the task instruction predictions and a similar decline for the corrective instruction and recovery mode prediction.
In the other two prediction tasks, our model also outperformed the variation that excludes the center crop input and the variant that only uses the \gls{ce} loss without scaling.
Although the margin for recovery mode predictions was smaller, with an improvement of around 0.5-1\%, the increase in corrective instruction predictions performance was more pronounced. 
This is particularly evident in the F1 score, highlighting the \gls{hl} policy's ability to issue language corrections more consistently, achieving a 2-2.5\% improvement.
Overall, the \gls{hl} policy achieved approximately 95\% accuracy in identifying recovery modes and around 70\% accuracy in predicting corrective instructions, out of 18 possible motion classes (see {Supplementary Methods "\nameref{corrective_instructions}").
We provide additional information on these evaluations in Table~\ref{tab:performance_metrics}.

As a further study, we apply GPT-4o, a state-of-the-art general-purpose vision-language model, as the \gls{hl} policy for surgical task planning.
GPT-4o was provided with the current endoscope image and all task instructions it could issue to guide the robot (see Fig.~\ref{fig:gpt_prompt}).
GPT-4o shows shortcomings in domain-specific understanding in issuing the correct task instruction.
For example, it initially omitted the crucial step of ``grabbing gallbladder" and prematurely initiated the action ``clipping first clip left tube". 
Additionally, GPT-4o incorrectly prompted the go-back from clipping/cutting instructions before completing the task.
Thus, GPT-4o would not be able to guide the \gls{ll} policy through a full cholecystectomy procedure, since it was unable to issue the correct task instructions.

\def\customlinethickness{0.8}
\def\custommarkersize{0.6}
\def\customendmarkersize{2.0}

\newcommand{\thumbnail}[2]{
    \begin{tikzpicture}
        \node (name) {#1};
        \node[anchor=west] at (name.east) {\includegraphics[width=0.5cm]{#2}};
    \end{tikzpicture}
}
\pgfplotsset{
    myAxisStyle/.style={
    view={30}{20},
    unit vector ratio=1 1 1,
    scale=0.6,
    xticklabel style={font=\tiny},
    yticklabel style={font=\tiny},
    zticklabel style={font=\tiny},
    xtick distance=10,
    ytick distance=10,
    ztick distance=10,
    xlabel=x,
    ylabel=y,
    zlabel=z,
    z label style={at={(-0.3, 0.5)}, rotate=-90},
    legend style={
        overlay,
        at={(0.5,3.1)},
        anchor=south,
        legend columns=2,
        font=\small,
        }, 
    }
}
\begin{figure}[htbp]
\vspace{2cm}
\begin{tikzpicture}
\begin{axis}[
    name=firstClip,
    myAxisStyle
]
\input{trajectoryplots/plot_0}
\legend{};
\end{axis}

\begin{axis}[
    name=thirdClip,
    at=(firstClip.right of south east),
    anchor=left of south west,
    myAxisStyle
]
\input{trajectoryplots/plot_1}
\end{axis}

\begin{axis}[
    name=cut,
    at=(thirdClip.right of south east),
    anchor=left of south west,
    myAxisStyle
]
\input{trajectoryplots/plot_2}
\legend{};
\end{axis}

\node[above=2mm of firstClip, anchor=south] (FC img) {\includegraphics[width=5cm]{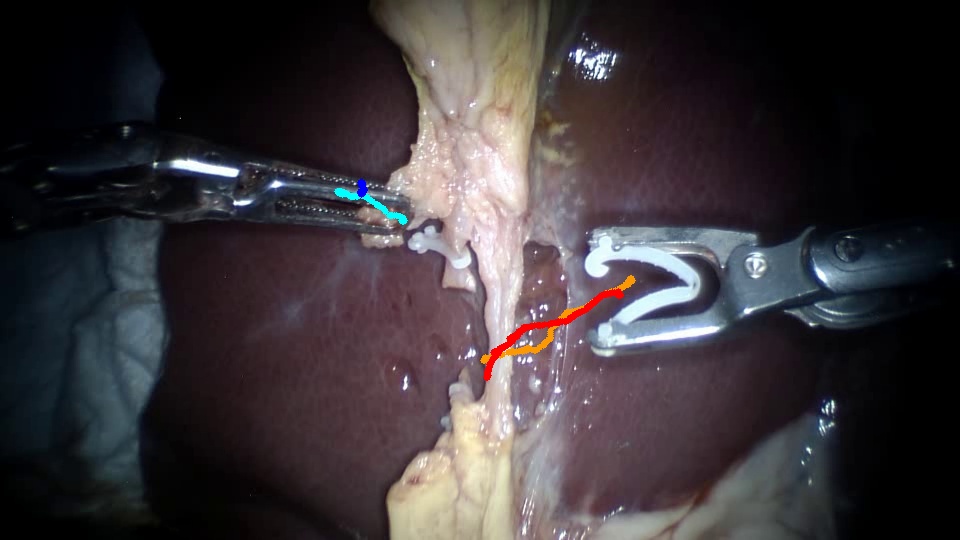}};
\node[above=2mm of thirdClip, anchor=south] (TC img) {\includegraphics[width=5cm]{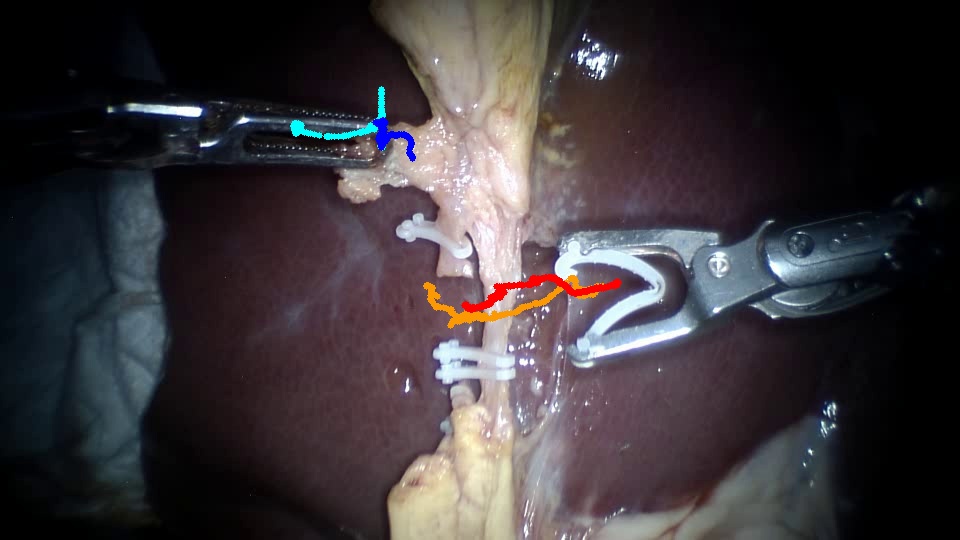}};
\node[above=2mm of cut, anchor=south] (C img) {\includegraphics[width=5cm]{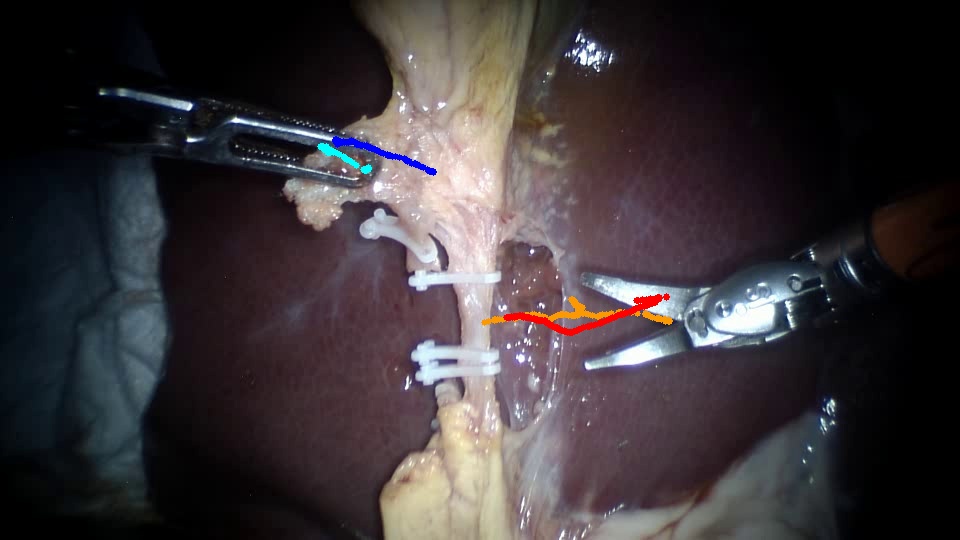}};

           \begin{axis}[
            xbar,
            at=(firstClip.south west),
            yshift=-0.7cm,
            xshift=8mm,
            anchor=north west,
            enlarge y limits=0.2,
            symbolic y coords={
                dur,
                len,
                jerk
            },
            ytick=data,
            ytick distance=-1, 
            width=0.33\textwidth,
            height=0.3\textwidth,
            x tick label style=transparent,
            y tick style=transparent,
            y tick label style={
                font=\footnotesize,
            },
            xmin=0,
            xmax=100,
            bar width=4mm,
            yticklabels={Duration (D) ($s$), Traj. Len. (TL) ($mm$), Mean Jerk (MJ) ($10^{-2}\frac{mm}{s^3}$)},
            y tick label style={
                rotate=45,
                anchor=east,
                inner sep=0pt,
                outer sep=1mm,
                align=center,
                font=\footnotesize,
            },
            y axis line style={draw=none,},
            axis x line=none,
            legend style={
                overlay,
                at={(0.0,-0.02)},
                anchor=north west,
                legend columns=2,
                font=\small,
                /tikz/every even column/.append style={column sep=1.5mm},
            reverse legend,
            },
            legend image code/.code={
                \draw [#1, draw=none] (0mm,-1mm) rectangle (1.5mm,2mm); },
            ]
            \addplot+ [
                fill=OIlightblue,
                draw=none,
            ] coordinates {
                (67,dur)
                (100,len)
                (100,jerk)
            };
           \addlegendentry{surgeon};

            \addplot+ [
                fill=OIblue,
                draw=none,
            ] coordinates {
                (100,dur)
                (58,len)
                (83,jerk)
            };
            \addlegendentry{SRT-H};

        \node[below=-0.5mm, anchor=north west, font=\footnotesize, text=black] at (axis cs:0,dur) {12};
        \node[above=-0.5mm, anchor=south west, font=\footnotesize, text=black] at (axis cs:0,dur) {18};
        
        \node[below=-0.5mm, anchor=north west, font=\footnotesize, text=black] at (axis cs:0,len) {54.005};
        \node[above=-0.5mm, anchor=south west, font=\footnotesize, text=black] at (axis cs:0,len) {31.387};
        
        \node[below=-0.5mm, anchor=north west, font=\footnotesize, text=black] at (axis cs:0,jerk) {3.058};
        \node[above=-0.5mm, anchor=south west, font=\footnotesize, text=black] at (axis cs:0,jerk) {2.544};
        
        \end{axis} 
           \begin{axis}[
            xbar,
            at=(thirdClip.south west),
            yshift=-0.7cm,
            xshift=5mm,
            anchor=north west,
            enlarge y limits=0.2,
            symbolic y coords={
                dur,
                len,
                jerk
            },
            ytick=data,
            ytick distance=-1, 
            yticklabels={,,},
            xmin=0,
            xmax=100,
            width=0.33\textwidth,
            height=0.3\textwidth,
            x tick label style=transparent,
            y tick style=transparent,
            y tick label style={
                font=\footnotesize,
            },
            bar width=4mm,
            yticklabels={D, TL, MJ},
            y tick label style={
                anchor=east,
                inner sep=0pt,
                outer sep=1mm,
                align=center,
                font=\footnotesize,
            },
            y axis line style={draw=none,},
            axis x line=none,
            ]

            \addplot+ [
                fill=OIlightblue,
                draw=none,
            ] coordinates {
                (76,dur)
                (100,len)
                (100,jerk)
            };
            
            \addplot+ [
                fill=OIblue,
                draw=none,
            ] coordinates {
                (100,dur)
                (69,len)
                (88,jerk)
            };
           
        \node[below=-0.5mm, anchor=north west, font=\footnotesize, text=black] at (axis cs:0,dur) {19};
        \node[above=-0.5mm, anchor=south west, font=\footnotesize, text=black] at (axis cs:0,dur) {25};
        
        \node[below=-0.5mm, anchor=north west, font=\footnotesize, text=black] at (axis cs:0,len) {63.072};
        \node[above=-0.5mm, anchor=south west, font=\footnotesize, text=black] at (axis cs:0,len) {43.450};
        
        \node[below=-0.5mm, anchor=north west, font=\footnotesize, text=black] at (axis cs:0,jerk) {3.050};
        \node[above=-0.5mm, anchor=south west, font=\footnotesize, text=black] at (axis cs:0,jerk) {2.675};
        \end{axis} 
           \begin{axis}[
            xbar,
            at=(cut.south west),
            yshift=-0.7cm,
            xshift=5mm,
            anchor=north west,
            enlarge y limits=0.2,
            symbolic y coords={
                dur,
                len,
                jerk
            },
            ytick=data,
            ytick distance=-1, 
            yticklabels={,,},
            xmin=0,
            xmax=100,
            width=0.33\textwidth,
            height=0.3\textwidth,
            x tick label style=transparent,
            y tick style=transparent,
            y tick label style={
                font=\footnotesize,
            },
            bar width=4mm,
            yticklabels={D, TL, MJ},
            y tick label style={
                anchor=east,
                inner sep=0pt,
                outer sep=1mm,
                align=center,
                font=\footnotesize,
            },
            y axis line style={draw=none,},
            axis x line=none,
            ]
            \addplot+ [
                fill=OIlightblue,
                draw=none,
            ] coordinates {
                (68,dur)
                (100,len)
                (100,jerk)
            };

            \addplot+ [
                fill=OIblue,
                draw=none,
            ] coordinates {
                (100,dur)
                (94,len)
                (89,jerk)
            };

        \node[below=-0.4mm, anchor=north west, font=\footnotesize, text=black] at (axis cs:0,dur) {13};
        \node[above=-0.4mm, anchor=south west, font=\footnotesize, text=black] at (axis cs:0,dur) {19};
        
        \node[below=-0.4mm, anchor=north west, font=\footnotesize, text=black] at (axis cs:0,len) {47.348};
        \node[above=-0.4mm, anchor=south west, font=\footnotesize, text=black] at (axis cs:0,len) {44.384};
        
        \node[below=-0.4mm, anchor=north west, font=\footnotesize, text=black] at (axis cs:0,jerk) {2.987};
        \node[above=-0.4mm, anchor=south west, font=\footnotesize, text=black] at (axis cs:0,jerk) {2.665};
        \end{axis} 

        \node[anchor=north east, xshift=-3mm] (A) at (FC img.north west) {\textbf{A}};
        \node[anchor=north, below = 3.5cm of A] (B) {\textbf{B}};
        \node[anchor=north, below = 3.25cm of B] {\textbf{C}};

\end{tikzpicture}
\caption{\textbf{Qualitative motion comparison between SRT-H and a surgeon}. We evaluate SRT-H against a human surgeon on the same gallbladder for the subtasks of applying the first and third clip to the artery, as well as cutting the artery. \textbf{(A)} 2D projection and \textbf{(B)} 3D plot of instrument paths for SRT-H (dark blue and red) and human surgeon (light blue and orange) as absolute positions in mm. \textbf{(C)} Quantitative comparisons between SRT-H and human surgeon based on the total duration of task execution (D in $s$) and trajectory length of the instruments (TL in $mm$), as well as the mean jerk (MJ in $10^{-2}\frac{mm}{s^3}$) calculated over the instrument paths.}
\label{fig:robot_vs_surgeon}
\end{figure}

\subsection*{\textbf{Comparison with expert surgeon}}
We perform a preliminary comparison between SRT-H and an expert surgeon.
Given the same gallbladder, both performed several tasks including adding the first and third clip to the artery and cutting it. Each round, SRT-H was deployed first and the surgeon was asked to repeat the same task. For adding the clips, modified clips with disabled latching mechanism were used. For cutting, right before the policy attempted to close its grippers to complete the cut, the robot was stopped to avoid permanent damage to the tissue. The surgeon had experience in performing both robotic and manual cholecystectomy. The surgeon did not have prior experience with the \gls{dvrk} system but was given sufficient time to become familiar with using the system. Note that the participating surgeon  study did not contribute to the training data.

The results are shown in Fig.~\ref{fig:robot_vs_surgeon}, which shows qualitative comparisons of the trajectories from the endoscope view and also in Cartesian space. We quantitatively report the mean jerk, trajectory length, and total duration during the tasks for both the surgeon and SRT-H. In general, we regard the better performer as the one that performs with the least mean jerk, trajectory length, and total duration. 

Our results show that the surgeon completes all tasks faster than SRT-H. However, we observed that SRT-H navigated with shorter trajectory length and less mean jerk compared to the surgeon, therefore SRT-H generates smoother and shorter trajectories. However, the surgeon was much faster in executing all the steps.
As a qualitative comparison, the 2D projections of the trajectories show that SRT-H and surgeon perform the procedure in a similar manner, based on the overall shape and appearance of the trajectories.
In general, despite these promising findings, we avoid making strong claims that SRT-H outperforms the surgeon. We also lacked a sufficient number of gallbladders for a more in-depth comparison. A more detailed analysis may be addressed in further extension of this work.
Our goal is to give an initial intuition of how our framework's performance compares to that of an experienced surgeon.

\section*{\textbf{DISCUSSION}}
In this work, we introduce SRT-H, a scalable framework for achieving step-level autonomy in robotic surgery. In comparison to prior work, which primarily focused on assistive tools~\cite{marescaux2003zeus, price2023using} and task-level autonomy~\cite{kuntz2023autonomous, shademan2016supervised, saeidi2022autonomous}, our research takes a step forward by moving toward autonomy at the step level. The results of our study demonstrate the effectiveness of SRT-H in automating the clipping and cutting procedure of a cholecystectomy intervention. Ablative studies show the effectiveness of our hierarchical design, which incorporates \gls{hl} and \gls{ll} policies. This design also demonstrates the ability to generalize across unseen ex-vivo tissues and self-correct errors in real-time. We demonstrate our approach across eight gallbladders, achieving a 100\% operation success rate.

\subsection*{Prior work}

\subsubsection*{\textit{Levels of autonomy}} The \Gls{loa} in medical robots is categorized across distinct levels~\cite{haidegger2019autonomy}, ranging from pure teleoperation to full autonomy. \Gls{loa} 0 represents no autonomy, where the robot functions purely as a tool controlled by a human operator. \Gls{loa} I is defined by robot assistance, where the robot provides continuous control support, such as mechanical guidance or virtual constraints, but the human remains in full control. \Gls{loa} II refers to task autonomy, where robots autonomously perform specific tasks, like running sutures, initiated by human input via discrete control commands. \Gls{loa} III, conditional autonomy, allows the system to generate task strategies autonomously but requires the human operator to select among them or approve an autonomously selected strategy. Systems at \gls{loa} IV, classified as high autonomy, can make medical decisions independently but still require supervision by a qualified doctor. Finally, \gls{loa} V represents full autonomy, where the robot is capable of performing an entire procedure without any human intervention. 

\subsubsection*{\textit{Examples of high LoAs}} Higher levels of autonomy \gls{loa} (IV) have been achieved by a few systems. One such system is the CyberKnife~\cite{kurup2010cyberknife}, which autonomously performs radiosurgery for brain and spine tumors under human supervision. This system operates in highly structured environments, using non-invasive techniques where tissues are rigid and stable, reducing the complexity of automation.
Another \gls{loa} IV system is the Veebot~\cite{perry2013profile}, which autonomously performs blood sampling by identifying and selecting suitable veins.
These systems demonstrate progress in autonomous surgery, however, they operate under controlled conditions, and the gap between these systems toward achieving full autonomy in dynamic, soft tissue environments remains considerable. 

Our present SRT-H work falls in \gls{loa} IV, as it is capable of reliable and autonomous execution, while self-correcting its mistakes; note that these self-corrective instructions are generated by itself and not issued by the user of the system. However, our system is not failure-proof to out-of-distribution scenarios, therefore the surgeon should always oversee its operation. 

Additionally, we briefly mention further evolved definitions of \gls{loa}, which include Level of Environmental Complexity (LoEC) and Level of Task Complexity (LoTC) \cite{LoA_evolved_definitions}. According to these metrics, our work falls in LoEC IV and LoTC IV. Our work can be categorized into LoEC IV because soft and realistic tissues are involved, although without topological motion (e.g., breathing), which is the further requirement needed to reach LoEC V. In terms of LoTC, our work falls into category IV because we consider advanced surgical tasks that require spatial understanding of the scene, but the model lacks clinical and anatomical knowledge, which is the further requirement to reach LoTC V.

We also draw a direct comparison to a highly relevant prior work involving autonomous bowel anastomosis \cite{saeidi2022autonomous}. Although anastomosis may seem like a more technically demanding task, our work demonstrates a greater step forward in comparison. More specifically, in this earlier work, the procedure took place under highly controlled conditions: the bowels were scaffolded on a fixture, fluorescent markers were used for tracking, and a specialized needle-throwing device simplified suturing to a basic reach task. Even with these advantages, the system occasionally made errors that required manual surgeon intervention. Moreover, the prior approach relied on a hand-crafted state-machine with model-based planning, which lacks expressivity. 
By contrast, our present work requires no special fixtures, tracking markers, or specialized surgical devices. Instead, it employs imitation learning to acquire more sophisticated and adaptable manipulation skills, which are difficult to capture with purely hand-crafted methods. For example, our system can delicately maneuver through the narrow space between the duct and artery, place clips at appropriate locations, and execute precise cuts without harming nearby tissue, all of which would be challenging to program explicitly. Crucially, the model can self-correct during the procedure, reducing the need for human intervention at test time. Furthermore, our method is expressive and scalable: by gathering demonstration data from additional procedures, we can potentially apply the same approach to a wide variety of surgical tasks, including anastomosis.

\subsubsection*{\textit{Robot transformers}} Outside of surgery, advancements in robotics have led to the development of general-purpose task-solving models~\cite{reed2022generalist, brohan2022rt, brohan2023rt, open_x_embodiment_rt_x_2023, hu2023Toward}.
These models are trained by imitation on extensive real-world robotics datasets, processing images from robot cameras, and following natural language task descriptions to generate robotic actions.
The resulting controllers exhibit the ability to adapt to novel situations and demonstrate task-solving capabilities that extend well beyond the scope of their training data~\cite{brohan2023rt}.
These models interpret commands that were not part of the training data and exhibit the ability to reason based on user instructions, such as which object to use as an improvised hammer (a rock) or finding a drink that is best for someone who is tired (an energy drink). 



\subsection*{Limitations}

\subsubsection*{\textit{From ex-vivo to in-vivo}} 

One important area for further research is translating our system from ex-vivo experiments to in-vivo clinical environments. Translating from ex-vivo to in-vivo brings several challenges, such as operating in the surgical site, addressing bleeding and tissue motion, and fitting the wrist cameras through laparoscopic ports. Since our approach is robot agnostic, and only depends on the relative position of the robot end-effectors, surgical access and operation do not present many challenges. Since our approach operates through visual guidance (instead of a model-based approach) and has the ability to self-correct, we believe it can adapt to motion and blood if it is incorporated as part of the training data or potentially zero-shot (see Fig.~\ref{fig:porcine_v_human} for reference). However, further studies are required to confirm this. Additionally, although the current wrist-camera configuration in our work would likely not fit into laparoscopic ports, modern cameras provide strong imaging quality with sub-millimeter form factors~\cite{ballester2024single, sayers2010optics} and can be easily integrated into surgical tools with minimal size increase of ports. Another concern with the use of wrist cameras may be potential occlusions due to fog and blood on the camera lenses. A potential solution to deal with these issues is to translate the strategies used for endoscopic cameras to wrist cameras. For instance, Anti-fogging solutions like Fred \cite{nezhat2008lensfogging} may be used for fogging scenarios. For blood wiping, there are commercial solutions like ClickClean \cite{clickclean2025} or ClearCam \cite{clearcam2025}, which physically remove any occlusions on the lens without removing the surgical tools. Furthermore, normalizing the usage of wrist cameras in the operating room may take time, considering they are devices not widely available in the market.

\subsubsection*{\textit{Making SRT-H safer}} 

A further extension fo this work may focus on expanding the system’s capabilities to cover a broader range of surgical procedures. The presented SRT-H framework supports the ability to learn across multiple surgical procedures using the same model parameters, to which diverse learning is believed to improve performance on individual tasks~\cite{brohan2022rt, brohan2023rt, schmidgall2024general}. 
Risk management remains a crucial aspect of surgical robotics. Further research could incorporate conservative Q-learning~\cite{chebotar2023q} and conformal prediction~\cite{ren2023robots, angelopoulos2021gentle} into the SRT-H system to address uncertainty during surgery. Conservative Q-Learning (CQL) would help prevent overestimation of the SRT-H's actions in unfamiliar situations, and conformal prediction would provide real-time feedback on the system’s confidence levels. Safety switching with robotic systems can be performed with on-site surgeons or through teleoperation, much like the proposed safety protocols used in autonomous driving systems~\cite{zhang2020toward, lim2023authority}. 
Additionally, with enhanced perception, it may be possible to simulate robot behaviors in simulation and refine plans before executing in the real-world for greater safety \cite{Gupta}.
Finally, although we demonstrate this approach primarily through full autonomy without supervision, our approach also supports real-time language interventions from expert surgeons, making it practical for potential integration into hospitals as a tool for surgeons to reduce fatigue on simple procedures or for areas with no access to trained surgeons. Intervention could be requested by the system based on uncertainty calculations and could be performed by a remote operator~\cite{ren2023robots}.

\section*{MATERIALS AND METHODS} \label{section:methods}

\subsection*{Data collection}

Training data was collected by two experienced human demonstrators on the \gls{dvrk} system. Dataset $D_1$, collected by the first demonstrator, contains data from 31 different gallbladders. 
The second demonstrator collected data for 3 additional gallbladders, which is denoted as dataset $D_2$. 
All gallbladder organs were sourced from Animal Technologies, Inc. (Tyler, TX, USA).
Note that both data collectors were non-clinical research assistants, trained by a surgical resident with extensive experience performing cholecystectomies. The first assistant was the primary data collector and contributed the most to the dataset. By the time the second data collector joined the project, most of the necessary data was collected therefore the contributed dataset was much smaller.
We define $D = D_1 \bigcup D_2$ as the union of both datasets. The visual data includes video streams from the \gls{dvrk} stereo endoscope, which has a resolution of $960 \times 540$ pixels, and two wrist cameras, each with a resolution of $640 \times 480$, mounted on the instruments of the surgical robot's left and right arm. Both video and kinematic streams are recorded at 30 \gls{fps}.

Prior to collecting specific task data, a demonstrator performed blunt dissection with Maryland forceps on a given gallbladder in order to reach the \gls{cvs}, where the cystic duct and artery are clearly identifiable. Certain gallbladders with abnormal tissue structures were not used, including the ones where the artery crosses over the duct and where the artery branches (see Fig. \ref{fig:anomalous_anatomy} for reference). 
Note that approximately 10\% of gallbladders were excluded because of these anatomical anomalies. Although the model can handle such variations if sufficient demonstration data are available, their rarity made it difficult to collect data at scale. Addressing these edge cases through scaled data collection is beyond the scope of this work and is left for future investigation. To simulate an accurate setup for the surgery, an expert surgeon recommended cholecystectomy port locations using a plastic abdominal dome. These ports were then isolated and modeled in computer-aided design (CAD) to create an open structure that holds the port locations for each arm of the surgical robot, as shown in Fig~\ref{fig:setup}A.  This way, the dissection area remains open rather than concealed, which is ideal for frequent wrist camera mounting, clip reloading, and tool switching. This open setup may raise concerns that ambient lighting may effect the lighting conditions. However, we found that its effect on the endoscopic and wrist cameras' image quality is negligible.

The clipping and cutting portions of cholecystectomy include 17 tasks in total. These include grabbing the gallbladder (1), adding six clips ($2 \times 6 = 12$), and cutting twice for the duct and artery ($2 \times 2 = 4$), summing to 17 ($1 + 12 + 4 = 17$). Note that the tasks for adding the clips and cutting involve two tasks: the motion for adding the clip or cutting and the retraction.  

In order to acquire multiple trials from a single gallbladder, we utilize a few tricks. For clipping motions, we use clips with the latching mechanism disabled. This allows us to perform clipping motions repetitively without actually locking it to the duct or the artery. For the cutting motions, we performed the motion of placing the scissors, but we do not close the scissors at the last step. During post-processing, we extend the kinematics data to simulate cutting motion. Using this strategy, it is possible to acquire multiple demonstration data using a single gallbladder with minimal damage. This may raise concerns that simply closing the grippers might not guarantee cutting. In practice, if the cut is not successful, which was very rare in our experiments, the policy often tried to cut again because it observed that the duct / artery was not cut and remained intact in the image observation. Also, multiple cuts were generally not necessary because the scissors were very sharp. We note that these strategies simply serve to aid with data collection without harming the tissues and do not take away from the generality of the methods.

We use the above logistics to collect many expert demonstrations. Additionally, we further collect samples that show recovery from suboptimal states to augment the dataset. These recovery demonstrations help the learned policies to recover from its own mistakes. 

After training the policies on the base dataset, we additionally collected a \gls{dagger}~\cite{kelly2019hg} dataset $D_{corr}$ as described in~\cite{shi2024yellrobotimprovingonthefly} to improve the base model performance by learning from verbal corrections
of common mistakes during policy rollout.
The \gls{dagger} algorithm iteratively collects data from the policy’s own actions and corrects them using expert feedback to refine the policy.
Within our \gls{dagger} dataset, only the language predictions are corrected, therefore it is denoted as HL DAgger for the rest of the paper.
The language corrections were either issued during the experiment or added during postprocessing.
The dataset is summarized in Table~\ref{tab:dataset_summary}, providing relevant statistics such as the number of demonstrations, images, and duration for both optimal and recovery demonstrations. These numbers represent the total number of trajectories collected across all gallbladders, encompassing all tasks involved in the clipping and cutting steps of the cholecystectomy procedure.

\begin{table}[htbp]
\caption{\textbf{Dataset summary.} 
\label{tab:dataset_summary}
Statistics for the data collected by the two main data collectors ($D_1$ \& $D_2$) and in the HL DAgger experiments ($D_{corr}$).}
\centering
\begin{tabular}{lcccc}
\toprule
 & & \textbf{Data Collector 1} & \textbf{Data Collector 2} & \textbf{HL DAgger} \\ 
\midrule
\textbf{Number of Gallbladders} & & 31 & 3 & 15 \\ 
\midrule
\multirow{3}{*}{\textbf{Optimal Demonstrations}} 
 & Num. &  12,304 &  885 & 264 \\ 
 & Images &  1,472,551 &  127,325 & 54,638 \\ 
 & Time (s) &  49,085 &  4,211 & 1,821 \\ 
\midrule
\multirow{3}{*}{\textbf{Recovery Demonstrations}} 
 & Num. &  4,904 &  263 & 352 \\ 
 & Images &  704,797 &  40,297 & 75,017 \\ 
 & Time (s) &  23,493 &  1,343 & 2,500 \\ 
\bottomrule
\end{tabular}
\end{table}

\subsection*{High-level policy}


\subsubsection*{\textit{Problem definition}}
The \gls{hl} policy, denoted $\pi_{\text{HL}}(p_t, c_t, m_t \mid o_{t-k:t})$, takes as input the current image observation $o$ at timestep $t$, along with $k$ preceding observations from the left camera stream of the \gls{dvrk} Si endoscope.
As output, the \gls{hl} policy generates three predictions: the next task $p_t$ (i.e. surgical phase) to be executed by the \gls{ll} policy, a correction flag $c_t$ indicating whether the robot is in a recovery mode, and a corrective (motion) instruction $m_t$ that specifies cardinal actions such as ``move right arm to the right" or ``move left arm towards me", which should be executed instead if the robot is in recovery mode.
A \gls{ce} loss is used for all three predicted outputs (see Eq.~\ref{eq:hl_loss_function}).
For the task instruction component, the \gls{ce} loss is scaled by the $L_1$ distance between the predicted and reference label to improve the policy's ability to distinguish between tasks that are temporally distant but visually similar.
The individual loss components are weighted based on their relative importance to the task.
The task instruction has the highest priority, so its weight $w_p = 0.4$ is set higher than the weight for the correction flag and corrective instruction predictions, which are set at $w_c=w_m=0.3$.
The resulting objective function minimizes the expected weighted sum of the task, correction, and motion losses and is given as the following. We use a hat symbol to denote the outputs predicted by the HL policy, while the corresponding ground-truth values from the dataset are written without the hat.
\begin{equation} \label{eq:hl_loss_function}
\begin{aligned}
\min_{\pi_{\text{HL}}} \mathbb{E}_{(o_{t-k:t}, p_t, c_t, m_t) \sim D} \bigg[ \ &  
    w_p \cdot \underbrace{L_{\text{CE}}\left(\pi_{\text{HL}}(\hat{p}_t \mid o_{t-k:t}), p_t\right)}_{\text{Task CE Loss}} \cdot 
    \underbrace{\left\Vert \hat{p}_t - p_t \right\Vert_1}_{\text{Task $L_1$ Distance}} \\
    & + w_c \cdot \underbrace{L_{\text{CE}}\left(\pi_{\text{HL}}(\hat{c}_t \mid o_{t-k:t}), c_t\right)}_{\text{Correction CE Loss}} 
    + w_m \cdot \underbrace{L_{\text{CE}}\left(\pi_{\text{HL}}(\hat{m}_t \mid o_{t-k:t}), m_t\right)}_{\text{Motion CE Loss}} 
\bigg].
\end{aligned}
\end{equation}


\subsubsection*{\textit{Model architecture}}
The \gls{hl} policy architecture, illustrated in Fig.~\ref{fig:hl_ll}~B, consists of a vision encoder, a Transformer Decoder~\cite{vaswani2017attentionneed}, and separate \gls{mlp} heads to generate the three classification outputs. 
Each image undergoes preprocessing, including standardization based on the mean and standard deviation of the color channels calculated over the entire dataset, ensuring zero mean and unit standard deviation. 
The image is resized to $224 \times 224$ to match the resolution used for pretraining the vision encoder. 
Alongside this global view, a centered crop that captures the most task-critical region is extracted and resized to $224 \times 224$. 
The centered crop covers the inner 50\% of the width and captures the lower 80\% of the height, starting from the bottom.
This approach is inspired by LLaVA's AnyRes technique~\cite{liu2024llavanext}, which divides images into multiple patches while preserving the global scene context. 
However, instead of generating multiple patches, we focus on extracting only the most task-relevant patch, emphasizing the center of the surgical area.
The vision encoder is the tiny variant of the Swin Transformer~\cite{Liu_2021_ICCV} pretrained on Imagenet~\cite{deng2009imagenet}.
The Swin Transformer is selected due to its high performance on limited data and its ability to produce a compact output token size of 768, which makes it suitable for temporal modeling with a downstream Transformer architecture.
During surgery, important details are often occluded.
For instance, a clip could easily be occluded by an instrument. 
In order to retain information crucial for classification, we include a history of $k=4$ past image frames, each spaced \SI{1}{\s} apart, along with the current frame as input to the \gls{hl} policy, following the approach of Shi et al.~\cite{shi2024yellrobotimprovingonthefly}. 
The embeddings from the vision encoder are used as inputs to the Transformer Decoder, configured with eight heads and six layers.
To preserve temporal information, sinusoidal position embeddings are added to the input sequence.
The vision encoder outputs are passed to the Transformer directly without pooling to preserve spatial information, similar to the approach by Zhao et al.~\cite{zhao2023learning}.
By assigning unique learnable embeddings as task-specific queries~\cite{zhao2023learning}, the Transformer Decoder can effectively attend to relevant spatial and temporal details, optimizing the alignment of each output with the most appropriate image frames and their features.

\subsubsection*{\textit{Training}}
The \gls{hl} policy base model is trained on dataset $D$ with the AdamW~\cite{loshchilov2019decoupledweightdecayregularization} optimizer, a learning rate of $1e^{-5}$, and a weight decay of $5e^{-2}$.
To improve both convergence and generalization, an annealing cosine weight schedule with a linear warmup of five epochs is applied.
We also incorporate data augmentation techniques, including RandAugment~\cite{cubuk2020randaugment} and coarse dropout from Albumentations~\cite{buslaev2020albumentations}, in order to boost visual robustness. 
Due to our specific dataset design, which consists of multiple individual recordings per task rather than continuous procedure recordings, two randomly sampled continuous task recordings are concatenated to artificially generate task transitions during \gls{hl} policy training.
To encourage the policy to learn a wider range of task semantics, 60\% of the input sequences begin within a recovery mode demonstration, exposing the policy to varying task executions and recovery scenarios. 
During training, we apply a prediction offset, where the policy is trained to predict the task instruction \SI{0.5}{\s} into the future rather than predicting the current surgical state.
This encourages the policy to anticipate upcoming actions and better handle task transitions~\cite{shi2024yellrobotimprovingonthefly}.
After the \gls{hl} \gls{dagger} dataset $D_{\text{corr}}$ is collected, the \gls{hl} policy is fine-tuned on the merged dataset $D \bigcup D_{corr}$.



\subsubsection*{\textit{Inference}}
Every \SI{3}{\s}, the \gls{hl} policy predicts a new task instruction $p_t$, correction flag $c_t$, and corrective instruction $m_t$.
Based on the corrective flag $c_t$, the language instruction provided to the \gls{ll} policy is then either the task instruction $p_t$ or the corrective motion $m_t$, as defined by Eq. \ref{switch_eq}:
\begin{equation}
l_t = 
\begin{cases} 
p_t, & \text{if } c_t = 0 \\
m_t, & \text{if } c_t = 1
\end{cases}
\label{switch_eq}
\end{equation}.

During inference, a human supervisor can override the \gls{hl} policy’s outputs via voice command or by selecting a task instruction or correction from a drop-down menu in our application \gls{gui}.
If a manual correction is made, the \gls{hl} policy outputs are overridden for the following \SI{3}{\s}.


\subsection*{Low-level policy}

\subsubsection*{\textit{Problem definition}}
The \gls{ll} policy is formulated as a language-conditioned policy $\pi_{\text{LL}}(a_{t:t+k} \mid o_t, l_t)$ to predict a sequence of robot actions $a_{t:t+k}$ based on the current image observation $o_t$ and language instruction $l_t$. $l_t$ can either be $p_t$ or $m_t$ depending on the correction flag $c_t$.
The input observations include the stereo endoscope's left image, along with images from the left and right wrist cameras.
For the action representation, we adopt the hybrid-relative action representation from~\cite{kim2024surgical}, which models relative Cartesian translations with respect to the endoscope tip and rotations relative to the end-effector.
This formulation compensates for the \gls{dvrk}'s kinematic inconsistencies~\cite{hwang2020calibrating}, leading to more consistent multi-step motion predictions. 
The policy is trained using behavior cloning, where the objective is to minimize the $L_1$ loss between the predicted action sequence and reference actions. The objective function is expressed in Eq. \ref{ll_equation}:
\begin{equation}
\begin{aligned}
\min_{\pi_{\text{LL}}} \mathbb{E}_{(o_t, l_t, a_{t:t+k}) \sim D} \left[ 
     \left\Vert\pi_{\text{LL}}(\hat{a}_{t:t+k} \mid o_t, l_t) - a_{t:t+k}\right\Vert_1\right].
\end{aligned} \label{ll_equation}
\end{equation}

\subsubsection*{\textit{Model architecture}}
The \gls{ll} policy is built on a decoder-only, BERT-like Transformer~\cite{devlin2019bert} that maps visual inputs to robot actions, as shown in Fig.~\ref{fig:hl_ll}B.
The visual inputs consist of images from the endoscope and wrist cameras, and they are encoded via a pre-trained EfficientNet-B3~\cite{tan2019efficientnet}. The encodings are then fused with language instruction embeddings from the \gls{hl} policy using feature-wise linear modulation (FiLM) layers.~\cite{perez2018film}. Language instructions are encoded using distilled bidirectional encoder representations from transformers (DistilBERT)~\cite{sanh2019distilbert}.
The fused visual and language embeddings, along with positional embeddings, are passed into the Transformer Decoder.
The action space is a 20-dimensional vector representing the relative actions for both robot arms (three for translation, six for rotation, and one for jaw angle per arm). Note that for rotation, we are using the 6D rotation formulation \cite{kim2024surgical,6d_rot}, where the rotation is represented by the first two columns of the rotation matrix. Its third column can be extrapolated by multiplying the first two columns, thus recovering the full rotation matrix. The six dimension rotation was shown to be more continuous than other rotation representations and thus easier for neural networks to learn.

With action chunking, the decoder outputs a $k \times 20$ tensor given the current observation. To optimize performance~\cite{zhao2023learning}, we predict robot actions for a 2 $s$ horizon, resulting in a chunk size of $k=60$.


\subsubsection*{\textit{Training}}


During training, the input images were resized to $224 \times 224$ pixels. To prevent overfitting, we apply several data augmentation techniques including random cropping, rotation, shifting, color jittering, and coarse dropout using Albumentations~\cite{buslaev2020albumentations}. 
Additionally, we apply a 7\% random dropout to one of the three input images, preventing the policy from over-relying on any single image observation.
To generate corrective language labels from the demonstration data, we examine a future chunk of actions and compute the motion trend along each axis. By comparing the magnitudes of motion across axes, we can determine the dominant direction of movement within that action segment. This enables assigning directional motion labels such as "move left arm to the right" or "move the right arm towards me." 
The chunk size here is set to 10 because we want to capture the unit of motions in the collected trajectories. If the chunk size is set too small, the generated instructions would be too noisy, and if the chunk size is too large, the more delicate motions would be ignored. 
During training, task instructions (e.g. ``grabbing gallbladder") are used when sampling from the base dataset, and corrective instructions (e.g., ``move left arm towards me") are used when sampling from recovery demonstrations. This enables the \gls{ll} policy to execute appropriate actions when given task instructions and recover from suboptimal states when given corrective instructions.
The policy contains approximately 72 million parameters and is trained on a single RTX 4090 GPU (24GB). Each epoch takes around \SI{4}{\min} with a batch size of 10, and training runs for 1500 epochs (\SI{100}{\hour}) before evaluation.

\subsubsection*{\textit{Inference}}
Inference time to produce a single action is approximately \SI{20}{\ms} on the same hardware.
To optimize performance, we set different execution horizons (the number of actions executed before resampling the \gls{ll} policy) for various phases of the procedure.
For the ``grabbing gallbladder" phase, we found through preliminary experiments that a shorter horizon caused the robot to change strategies too frequently, leading to hesitation and continuous pose adjustments without fully committing to a successful strategy. Setting the execution horizon to 30 timesteps ensures that the robot commits to a single strategy.
In contrast, for the other phases, we set a shorter execution horizon of 20 timesteps to enable more frequent replanning. These phases require high precision, particularly when maneuvering the right arm between the duct and artery.
Additionally, manual tool switching and clip loading between tasks were necessary during experiments. To manage these transitions, we implemented a logic-based state machine to automatically pause both the \gls{hl} and \gls{ll} policies during phase transitions. For instance, the pauses were triggered when shifting from ``going back from the first clip left tube" to ``clipping second clip left tube" or from ``going back from third clip right tube" to ``going to the cutting position on the right tube".

\subsection*{Statistical analysis}

The mean computed in Fig. \ref{fig:variant_comparisons} with sample size $N$ and data point $x$ were computed using the following equation:
\[
\mu = \frac{1}{N} \sum_{i=1}^{N} x_i,
\]

\noindent

\bibliography{iclr2021_conference}
\bibliographystyle{iclr2021_conference}

\clearpage

\section*{ACKNOWLEDGEMENTS}

\paragraph*{Funding:}

Research reported in this publication was supported by the Advanced Research Projects Agency for Health (ARPA-H) under Award Number 75N91023C00048, as well as NSF/FRR 2144348, NIH R56EB033807, and NSF DGE 2139757. The views and conclusions contained in this document are those of the authors and should not be interpreted as representing the official policies, either expressed or implied, of the U.S. Government.


\paragraph*{Author contributions:}

Conceptualization: J.W.K., A.K., S.S., D.R.T, R.C., C.F.;
Methodology: J.W.K., C.F., A.K.;
Software: J.W.K., L.X.S., P.H., J.C., A.D., P.M.S.;
Visualization: J.W.K., S.S., P.H., J.C., P.M.S.;
Data Curation: A.G., J.W.K., P.H., J.J., B.W.;
Formal analysis: J.W.K., P.H., J.C.;
Funding acquisition: A.K., R.C.;
Supervision: A.K., J.W.K., C.F.; 
Writing—original draft: J.W.K., S.S., P.H., J.C., P.M.S., A.G.;
Writing—review and editing: J.W.K., S.S., P.H., J.C., P.M.S., A.G., A.K., C.F., L.X.S, D.R.T, R.C.;

\paragraph*{Competing interests:}
Provisional Patent Pending: "Imitation learning for surgical robots with kinematics errors using self-corrections." Richard Cha has ownership interests in and serves as a scientific advisor for Optosurgical, LLC.

\paragraph*{Data and materials availability:}

All data supporting the conclusions of this paper are included in the main text or Supplementary Materials. The datasets and code used to generate Fig. 4, 6, and S7 are available at Zenodo: https://zenodo.org/records/15637074

\subsection*{Supplementary materials}
Supplementary Methods \\
Figs. S1 to S7\\
Tables S1 to S2\\
References \textit{(7-\arabic{enumiv})}\\ 


\newpage


\renewcommand{\thefigure}{S\arabic{figure}}
\renewcommand{\thetable}{S\arabic{table}}
\renewcommand{\theequation}{S\arabic{equation}}
\renewcommand{\thepage}{S\arabic{page}}
\setcounter{figure}{0}
\setcounter{table}{0}
\setcounter{equation}{0}
\setcounter{page}{1} 


\section*{Supplementary Materials}



\subsubsection*{This PDF file includes:}
Supplementary Methods\\
Tables S1 to S2\\
Figures S1 to S7\\


\newpage

\setcounter{section}{0}
\renewcommand{\thesubsection}{\thesection.\arabic{subsection}}
\section*{Supplementary materials and methods}
\subsection*{High-level policy training configuration}
The policy consists of approximately 45 million parameters, with 29 million allocated to the encoder and the remaining 16 million shared between the Transformer and \gls{mlp} heads.
The training was conducted on a single RTX 4090 GPU (24GB), taking around 20 hours to complete 500 epochs (4,000 iterations each).
The inference time is approximately 25 ms.
During training, we used a 90-10 split for the training and validation sets and selected the model weights that performed best on the validation set.
For the \gls{hl} ablation study, shown in Table~\ref{tab:performance_metrics}, we sampled 40,000 sequences from the validation dataset and applied all variants on the same input sequences.
During training, we applied RandAugment~\cite{cubuk2020randaugment}, a data augmentation technique that automatically selects and applies a subset of augmentations from a predefined set of transformations (listed in Table~\ref{tab:randAugment}).

\subsection*{Corrective language instructions} \label{corrective_instructions}
The high-level policy was capable of generating 18 corrective instructions, including: close left gripper, close right gripper, open left gripper, open right gripper, move left arm to the left, move left arm to the right, move left arm towards me, move left arm away from me, move left arm higher, move left arm lower, move right arm to the left, move right arm to the right, move right arm towards me, move right arm away from me, move right arm higher, move right arm lower, close both grippers, open both grippers.

\subsection*{General purpose \gls{vlm} as high-level policy}
 an alternative to the selected \gls{hl} policy, we explore using a state-of-the-art general-purpose \gls{vlm}, GPT-4o, to perform the \gls{hl} task planning.
To evaluate GPT-4o's potential, we test it in the same setup as our \gls{hl} policy, assigning it the role of a surgical task planner for the \gls{dvrk}.
GPT-4o is provided with the current endoscope image and the necessary task instructions it could issue to the robot, as shown in Fig.~\ref{fig:gpt_prompt}.
To provide more spatial context, clipping task instructions include additional information on where the clip should be placed (\eg, bottom or top), and filler instructions such as ``reload clip" or ``exchange instrument" are added.
Processing the first frame with GPT-4o already highlights the model's lack of domain-specific knowledge and its difficulties with visual recognition of task completion and transitions. 
For example, the model initially skips the critical step of ``Grabbing the Gallbladder" (Fig.~\ref{fig:gpt_t0}), only selecting this instruction after being prompted that the gallbladder had not been grabbed yet.
Similar errors are observed later; GPT-4o triggered the ``Clipping the Left Tube (Bottom Clip 1)" step prematurely, before successfully grabbing the gallbladder (Fig.~\ref{fig:gpt_t2}), and incorrectly prompts the transition to go back from clipping before the clip has been set (Fig.~\ref{fig:gpt_t5}).
These observations indicate that a general-purpose \gls{vlm} like GPT-4o lacks the task-specific precision required for effective surgical task planning.
Fine-tuning on this specific domain is required, as reliance on prompt engineering alone proves insufficient for our surgical task planner setup.

\subsection*{Supplementary Figures and Tables}

\begin{figure}[htbp]
    \centering
    \includegraphics[width=0.75\linewidth]{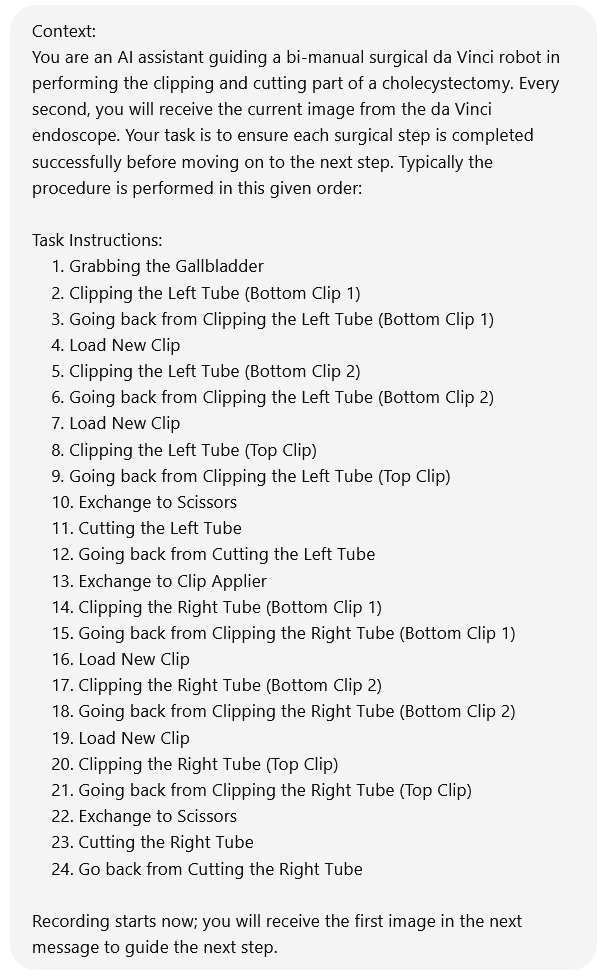}
    \caption{Prompt given to GPT-4o. Describing first the context including the role and which input data it will receive as well as the task instructions that it can later issue. We only provide endoscope images as input without wrist camera images, consistent with the input of the \gls{hl} policy.}
    \label{fig:gpt_prompt}
\end{figure}

\begin{figure}[htbp]
    \centering
    \includegraphics[width=0.6\linewidth]{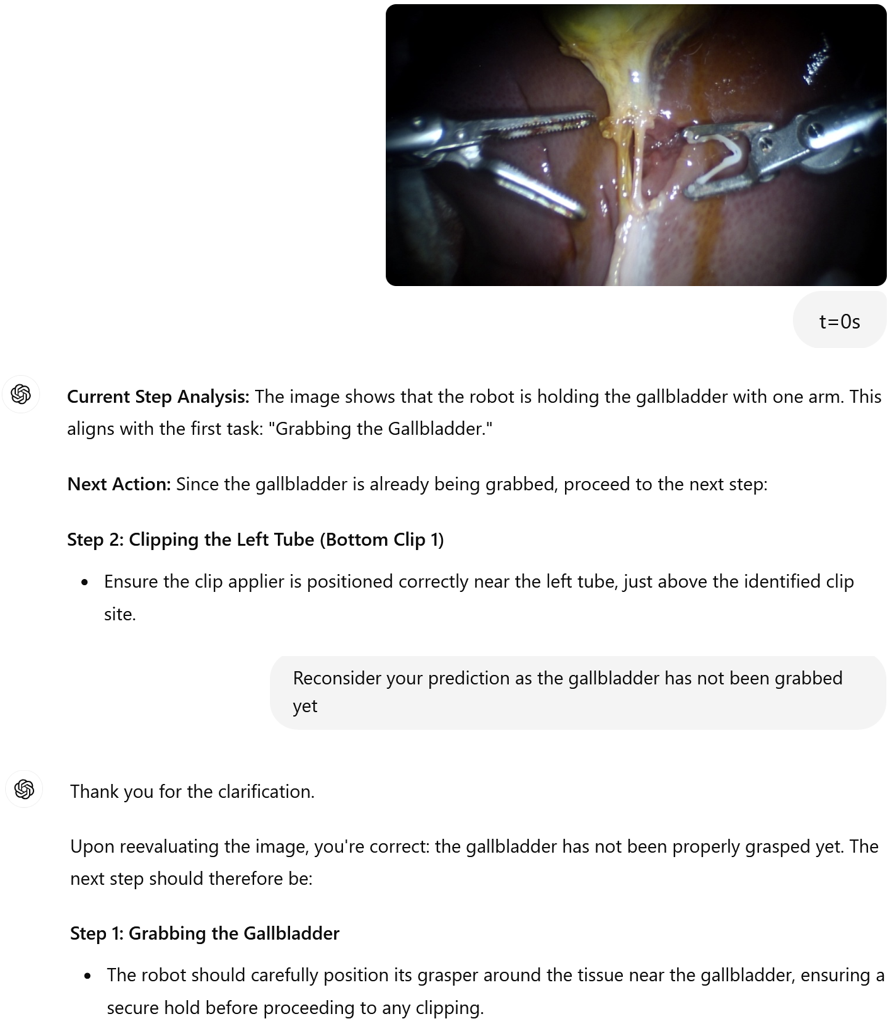}
    \caption{Response for input t=0s. Skipping the ``Grabbing the Gallbladder" surgical step.}
    \label{fig:gpt_t0}
\end{figure}

\begin{figure}[htbp]
    \centering
    \includegraphics[width=0.6\linewidth]{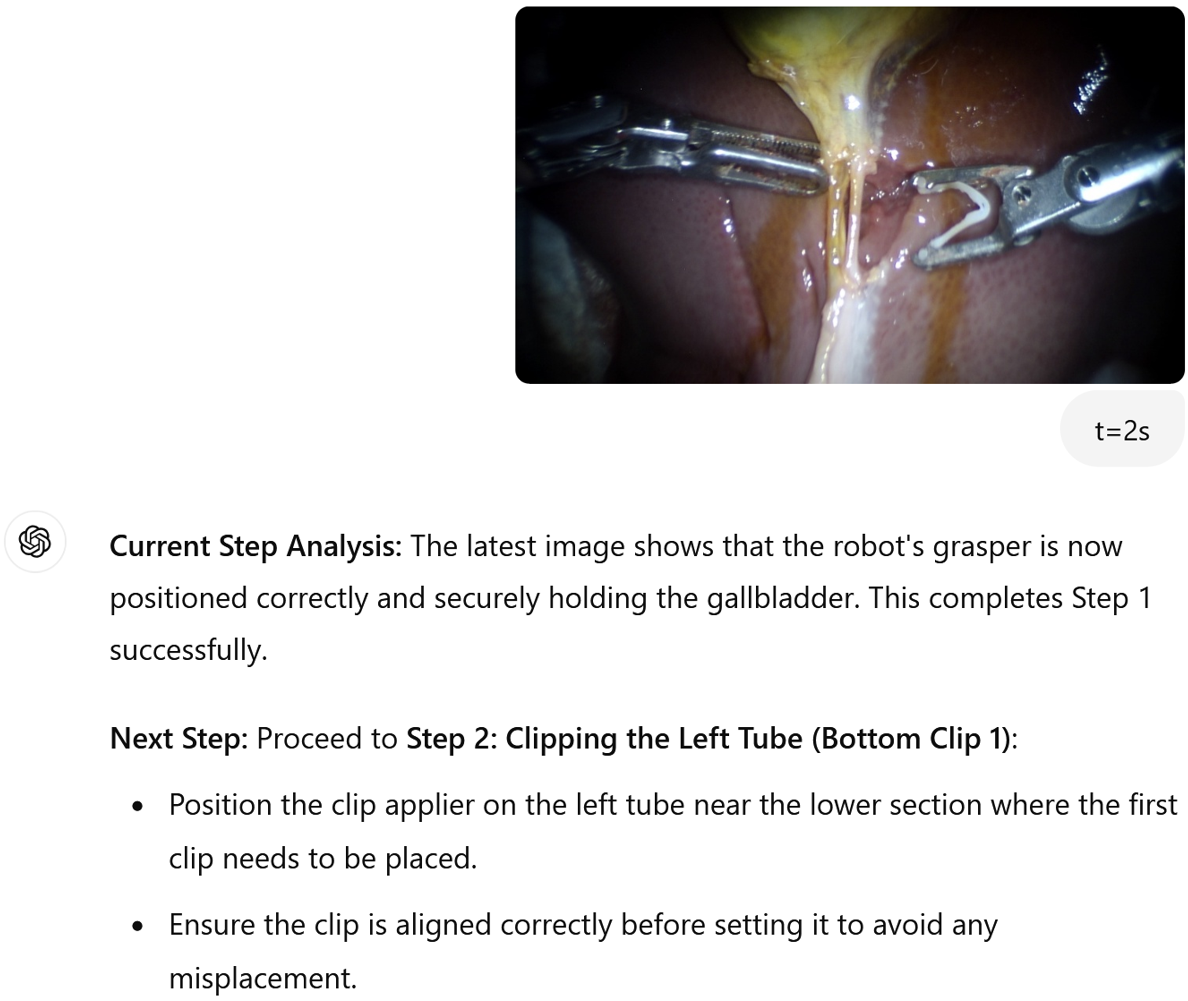}
    \caption{Response for input t=2s. Triggering too early to the clipping task instruction.}
    \label{fig:gpt_t2}
\end{figure}

\begin{figure}[htbp]
    \centering
    \includegraphics[width=0.6\linewidth]{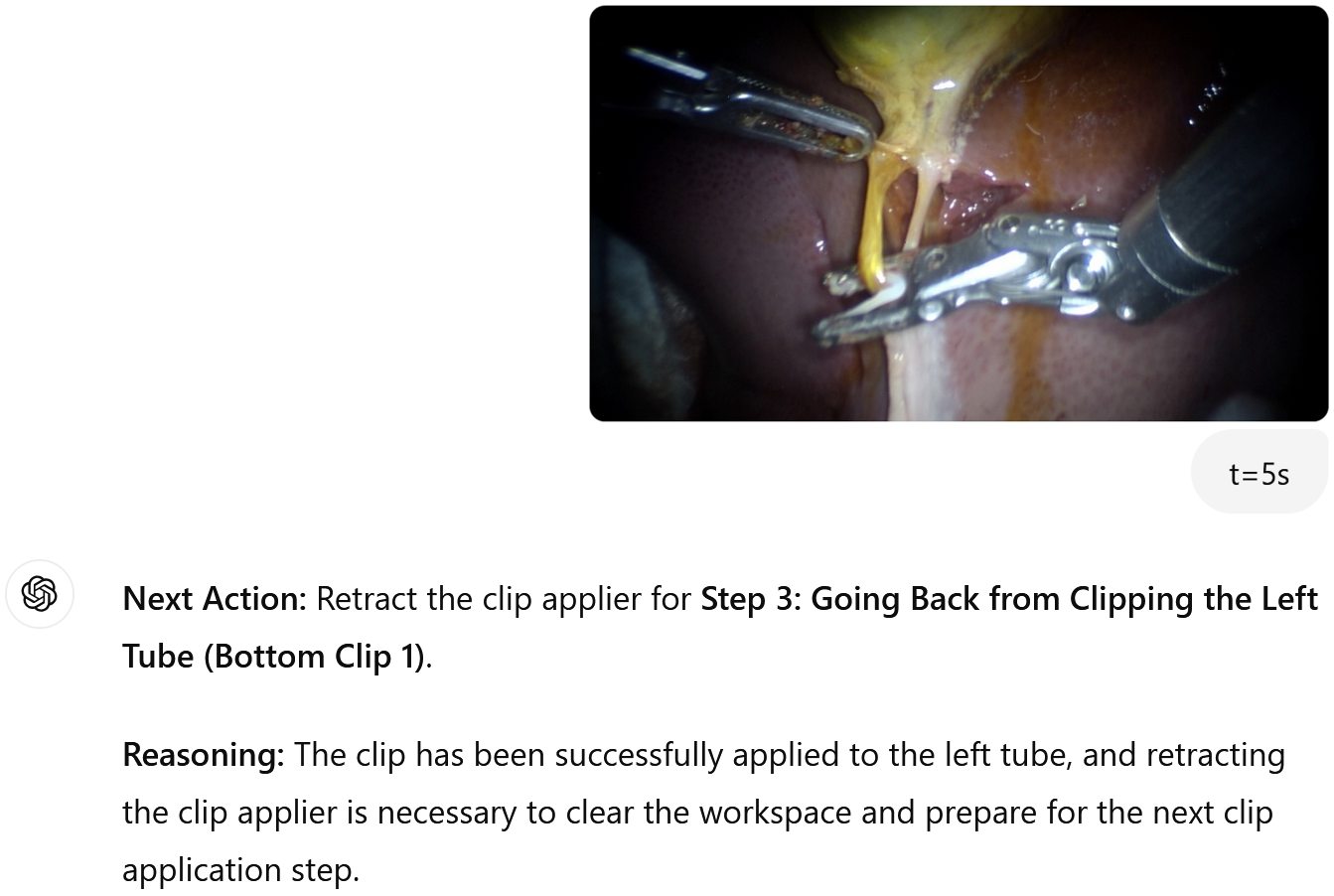}
    \caption{Response for input t=5s. Triggering too early to go back from clipping when the clip was not set yet.}
    \label{fig:gpt_t5}
\end{figure}

\begin{table}[htbp]
    \centering
    \caption{Image transformations used in RandAugment~\cite{cubuk2020randaugment}.}
    \label{tab:randAugment}
    \begin{tabular}{lll}
        \textbullet\ identity        & \textbullet\ autoContrast    & \textbullet\ equalize        \\
        \textbullet\ rotate          & \textbullet\ solarize        & \textbullet\ color           \\
        \textbullet\ posterize       & \textbullet\ contrast         & \textbullet\ brightness      \\
        \textbullet\ sharpness       & \textbullet\ shear-x          & \textbullet\ shear-y         \\
        \textbullet\ translate-x     & \textbullet\ translate-y     &                                \\
    \end{tabular}
\end{table}

\begin{figure}[htbp]
    \centering
    \includegraphics[width=0.9\linewidth]{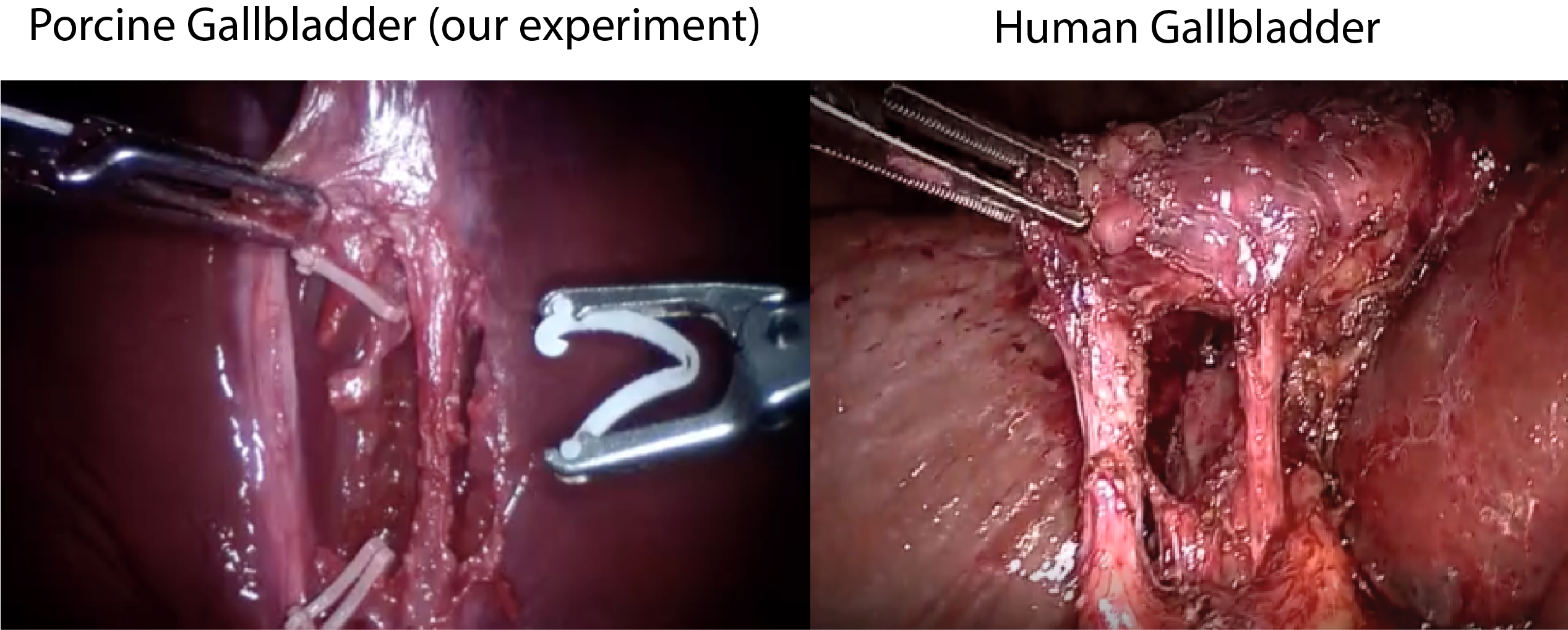}
    \caption{Difference between porcine (left) and human (right) gallbladder.}
    \label{fig:porcine_v_human}
\end{figure}

\begin{figure}[htbp]
    \centering
    \includegraphics[width=0.9\linewidth]{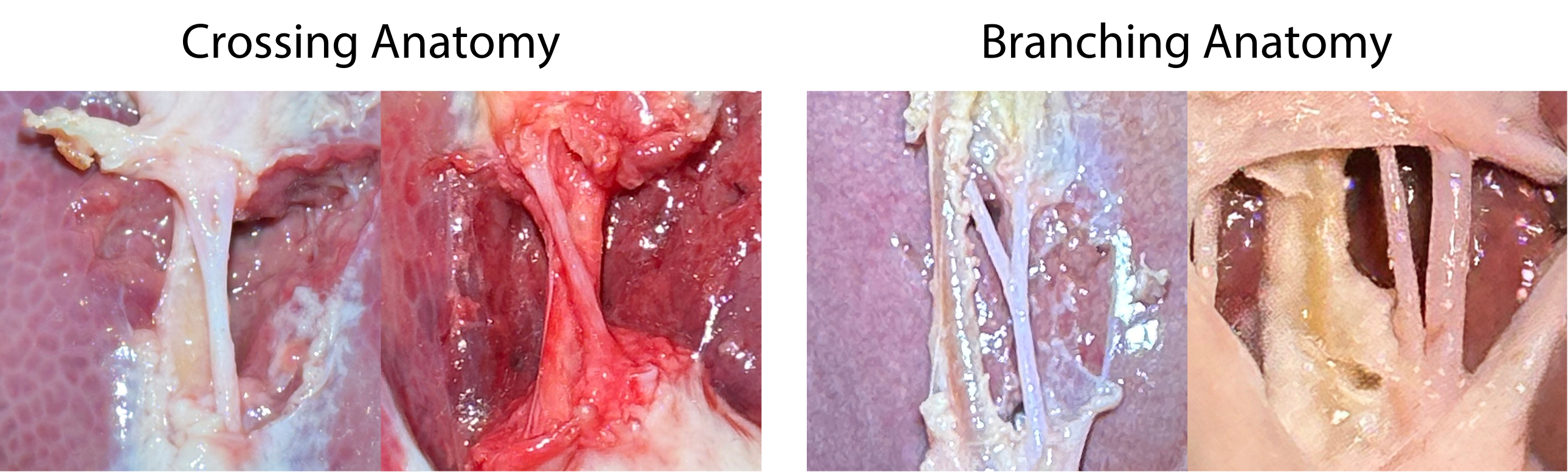}
    \caption{Examples of anomalous gallbladder anatomy.}
    \label{fig:anomalous_anatomy}
\end{figure}

\begin{figure}[htbp]
        \begin{tikzpicture}
        \begin{scope}[local bounding box=scope name]
            \def\angle{0}
            \def\radius{3.0}
            \def\cyclelist{{
            "OIblue", 
            "OIred", 
            "OIgreen", 
            "OIlightblue", 
            "OIpink", 
            "OIyellow", 
            "OIorange", 
            "OIgreen", 
            "OIblue",
            "OIred",
            "OIpink"
            }}
            \def\patternlist{{
            "none", 
            "none", 
            "none", 
            "none", 
            "none", 
            "none", 
            "none", 
            "dots", 
            "dots",
            "dots",
            "dots"
            }}
            \def\patterncolorlist{{
            "black", 
            "black", 
            "black", 
            "black", 
            "black", 
            "black", 
            "black", 
            "black", 
            "black",
            "black",
            "black"
            }}
            \newcount\cyclecount \cyclecount=-1
            \newcount\ind \ind=-1
          \foreach \occcount/\name in {
              11/move r. a. \textrightarrow,
              17/move r. a. \textleftarrow,
              7/close r. gripper,
              5/move r. a. $\odot$,
              2/move r. a $\otimes$,
              4/move r. a. \textdownarrow,
              1/move r. a. \textuparrow,
              1/open l. gripper,
              1/move l. a. \textrightarrow,
              1/move l. a. \textleftarrow,
              1/move l. a $\otimes$
            } {
              \ifx\occcount\empty\else               
                \global\advance\cyclecount by 1     
                \global\advance\ind by 1            
                \ifnum11<\cyclecount                 
                  \global\cyclecount=0              
                  \global\ind=0                     
                \fi
                \pgfmathparse{\cyclelist[\the\ind]} 
                \edef\wcolor{\pgfmathresult}         
                \pgfmathparse{\patternlist[\the\ind]} 
                \edef\pattern{\pgfmathresult}         
                \pgfmathparse{\patterncolorlist[\the\ind]} 
                \edef\pcolor{\pgfmathresult}         
                \pgfmathparse{divide(\occcount, 51)*100}
                \edef\percent{\pgfmathresult}         
                \draw[fill={\wcolor},draw={\wcolor}] (0, 0) -- (\angle:\radius)
                  arc (\angle:\angle+\percent*3.6:\radius) -- cycle;
                \draw[pattern=\pattern, pattern color=\pcolor] (0, 0) -- (\angle:\radius)
                  arc (\angle:\angle+\percent*3.6:\radius) -- cycle;
                \node[font=\footnotesize] at (\angle+0.5*\percent*3.6:0.8*\radius) {\textcolor{black}{\occcount}};
                \pgfmathparse{\angle+\percent*3.6}  
                \xdef\angle{\pgfmathresult}         
              \fi
            };
        \end{scope}

        \node[anchor=south] (pie title) at ($(scope name.north) + (0.3cm, 0.85cm)$) {Total \# Self Corrections (n=8)};

        \node (legend) at ($(scope name.east) + (0.6cm, 3cm)$) {};
        \node (A) [below = 0 of legend, anchor=north west, font=\footnotesize] {move r. a. \textrightarrow};
        \draw [fill=OIblue] (A.south west)+(-0.2,0.1) rectangle (A.north west)+(-0.1,0.0);
        
        \node (B) [below = 0 of A.south west, anchor=north west, font=\footnotesize] {move r. a. \textleftarrow};
        \draw [fill=OIred] (B.south west)+(-0.2,0.1) rectangle (B.north west)+(-0.1,0.0);

        \node (C) [below = 0 of B.south west, anchor=north west, font=\footnotesize] {close r. gripper};
        \draw [fill=OIgreen] (C.south west)+(-0.2,0.1) rectangle (C.north west)+(-0.1,0.0);

        \node (D) [below = 0 of C.south west, anchor=north west, font=\footnotesize] {move r. a. $\odot$};
        \draw [fill=OIlightblue] (D.south west)+(-0.2,0.1) rectangle (D.north west)+(-0.1,0.0);

        \node (E) [below = 0 of D.south west, anchor=north west, font=\footnotesize] {move r. a $\otimes$};
        \draw [fill=OIpink] (E.south west)+(-0.2,0.1) rectangle (E.north west)+(-0.1,0.0);

        \node (F) [below = 0 of E.south west, anchor=north west, font=\footnotesize] {move r. a. \textdownarrow};
        \draw [fill=OIyellow] (F.south west)+(-0.2,0.1) rectangle (F.north west)+(-0.1,0.0);

        \node (G) [below = 0 of F.south west, anchor=north west, font=\footnotesize] {move r. a. \textuparrow};
        \draw [fill=OIorange] (G.south west)+(-0.2,0.1) rectangle (G.north west)+(-0.1,0.0);

        \node (H) [below = 0 of G.south west, anchor=north west, font=\footnotesize] {open l. gripper};
        \draw [fill=OIgreen] (H.south west)+(-0.2,0.1) rectangle (H.north west)+(-0.1,0.0);
        \draw [pattern=dots] (H.south west)+(-0.2,0.1) rectangle (H.north west)+(-0.1,0.0);

        \node (I) [below = 0 of H.south west, anchor=north west, font=\footnotesize] {move l. a. \textrightarrow};
        \draw [fill=OIblue] (I.south west)+(-0.2,0.1) rectangle (I.north west)+(-0.1,0.0);
        \draw [pattern=dots] (I.south west)+(-0.2,0.1) rectangle (I.north west)+(-0.1,0.0);

        \node (J) [below = 0 of I.south west, anchor=north west, font=\footnotesize] {move l. a. \textleftarrow};
        \draw [fill=OIred] (J.south west)+(-0.2,0.1) rectangle (J.north west)+(-0.1,0.0);
        \draw [pattern=dots] (J.south west)+(-0.2,0.1) rectangle (J.north west)+(-0.1,0.0);

        \node (K) [below = 0 of J.south west, anchor=north west, font=\footnotesize] {move l. a $\otimes$};
        \draw [fill=OIpink] (K.south west)+(-0.2,0.1) rectangle (K.north west)+(-0.1,0.0);
        \draw [pattern=dots] (K.south west)+(-0.2,0.1) rectangle (K.north west)+(-0.1,0.0);

        \begin{axis}[
            name=SelfCorrections,
            at=(scope name.east),
            xshift=5cm,
            anchor=west,
            ybar stacked,
            nodes near coords,
            every node near coord/.append style={font=\scriptsize, text=black},
            title={Self Corrections per Subtask (n=8)},
            symbolic x coords={grasp, clip, cut, back},
            xticklabels={grab, clip, cut, go back},
            xtick=data,
            ymin=0,
            ymax=28,
            xlabel={},
            ylabel={\# Occurences},
            width=0.5\textwidth,
            height=0.55\textwidth,
            x tick label style={
                inner sep=0pt,
                outer sep=1mm,
                align=center,
                font=\footnotesize,
                yshift=-2.0mm,
            },
            y tick label style={
                font=\footnotesize,
            },
            xtick pos=bottom,
            bar width=8mm,
            ymajorgrids=true,
            yminorgrids=true,
            minor y tick num=2,
            major grid style={thick, black!30!white, dashed},
            minor grid style={ultra thin, black!20!white, dashed},
            major x tick style = transparent,
            minor y tick style = transparent,
            enlarge x limits=0.1,
            legend style={
                legend columns=3,
                font=\small,
                /tikz/every even column/.append style={column sep=1.5mm},
            },
            legend pos=north west,
            legend image code/.code={
                \draw [#1, draw=none] (0mm,-1mm) rectangle (1.5mm,2mm);
                },
            ]
            
            \addplot+[ybar, black, fill=OIblue, draw=none] plot coordinates {
                (grasp, 0)
                (clip, 3)
                (cut, 8)
                (back, 0)
            };
            \addplot+[ybar, black, fill=OIred, draw=none] plot coordinates {
                (grasp, 0)
                (clip, 5)
                (cut, 11)
                (back, 1)
            };
            \addplot+[ybar, black, fill=OIgreen, draw=none] plot coordinates {
                (grasp, 0)
                (clip, 3)
                (cut, 4)
                (back, 0)
            };
            \addplot+[ybar, black, fill=OIlightblue, draw=none] plot coordinates {
                (grasp, 0)
                (clip, 3)
                (cut, 2)
                (back, 0)
            };
            \addplot+[ybar, black, fill=OIpink, draw=none] plot coordinates {
                (grasp, 0)
                (clip, 1)
                (cut, 1)
                (back, 0)
            };
            \addplot+[ybar, black, fill=OIyellow, draw=none] plot coordinates {
                (grasp, 0)
                (clip, 3)
                (cut, 1)
                (back, 0)
            };
            \addplot+[ybar, black, fill=OIorange, draw=none] plot coordinates {
                (grasp, 0)
                (clip, 1)
                (cut, 0)
                (back, 0)
            };
            \addplot+[ybar, black, fill=OIgreen, draw=none, postaction={ pattern=dots}] plot coordinates {
                (grasp, 1)
                (clip, 0)
                (cut, 0)
                (back, 0)
            };
            \addplot+[ybar, black, fill=OIblue, draw=none, postaction={ pattern=dots}] plot coordinates {
                (grasp, 1)
                (clip, 0)
                (cut, 0)
                (back, 0)
            };
            \addplot+[ybar, black, fill=OIred, draw=none, postaction={ pattern=dots}] plot coordinates {
                (grasp, 1)
                (clip, 0)
                (cut, 0)
                (back, 0)
            };
            \addplot+[ybar, black, fill=OIpink, draw=none, postaction={ pattern=dots}] plot coordinates {
                (grasp, 1)
                (clip, 0)
                (cut, 0)
                (back, 0)
            };
            
        \end{axis}
        
        \node[anchor=east, xshift=-3mm] (B) at (SelfCorrections.north west) {\textbf{B}};
        \node[anchor=east, xshift=-10cm] (A) at (B.west) {\textbf{A}};
            
        \end{tikzpicture}
        
    \caption{\textbf{High-Level Language Policy corrections.} We analyse the occurrence of self corrections that were generated by the High-Level Language Policy to guide the Low-Level Policy in recovery modes over n=8 experiments. Left arm and right arm of the robot are abbreviated to r. a. and l. a., respectively. The cardinal directions are shown as symbols for right \textrightarrow, left \textleftarrow, higher \textuparrow, lower \textdownarrow, towards me $\odot$, away from me $\otimes$. \textbf{(A)} shows the total number of self corrections over all experiments and \textbf{(B)} a detailed view for each correction type distributed over subtasks grouped into grab, clip, cut, and go back.}
    \label{fig:selfcorrections}
\end{figure}
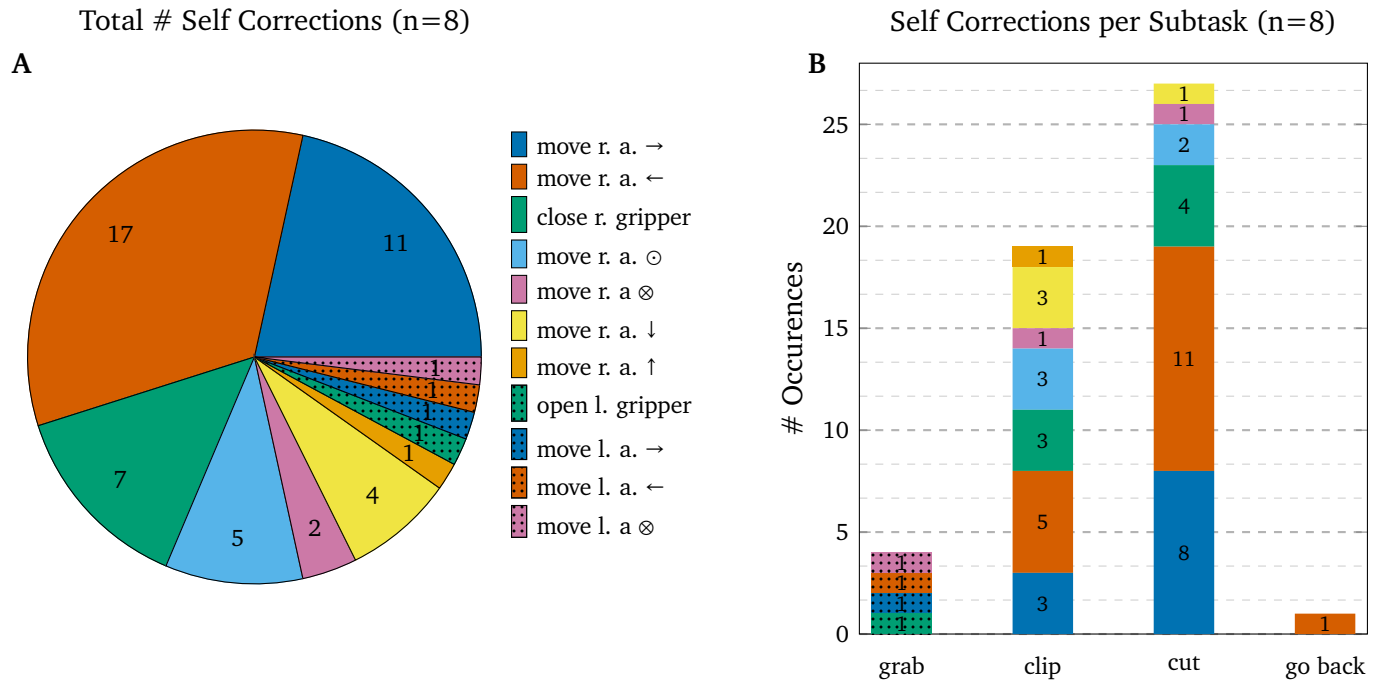

\begin{table}[htbp]
\caption{\textbf{\gls{hl} 
\label{tab:performance_metrics}Policy Performance Metrics.} Prediction accuracy and F1 score of the \gls{hl} policy on the individual targets for task instructions, correction flag, and corrective instruction.}
\centering
\begin{tabular}{lccccc}
\toprule
 & & \textbf{Ours} & \textbf{No Center Crop} & \textbf{No $L_1$ Dist.} & \textbf{No Hist.} \\ 
\midrule
\multirow{2}{*}{\textbf{Task Instruction}} 
 & Accuracy (\%) \textuparrow & \textbf{97.29} & 94.80 & 94.99 & 85.44 \\ 
 & F1 Score (\%) \textuparrow & \textbf{97.18} & 94.82 & 94.60 & 84.56 \\ 
\midrule
\multirow{2}{*}{\textbf{Correction Flag}} 
 & Accuracy (\%) \textuparrow & \textbf{95.48} & 94.70 & 95.15 & 87.55 \\ 
 & F1 Score (\%) \textuparrow & \textbf{91.03} & 89.25 & 90.35 & 57.07 \\ 
\midrule
\multirow{2}{*}{\textbf{Corrective Instruction}} 
 & Accuracy (\%) \textuparrow & \textbf{69.65} & 67.85 & 68.35 & 60.55 \\ 
 & F1 Score (\%) \textuparrow & \textbf{56.37} & 52.73 & 53.91 & 44.42 \\ 
\bottomrule
\end{tabular}
\end{table}

\clearpage 


\end{document}